\documentclass[journal]{IEEEtran}
\usepackage{amsmath,amsfonts}
\usepackage{algorithmic}
\usepackage{algorithm}
\usepackage{array}
\usepackage[caption=false,font=normalsize,labelfont=sf,textfont=sf]{subfig}
\usepackage{textcomp}
\usepackage{stfloats}
\usepackage{url}
\usepackage{verbatim}
\usepackage{graphicx}
\usepackage{cite}

\usepackage{tabularx}
\usepackage{makecell}
\usepackage{multirow}
\usepackage{booktabs}
\usepackage{graphicx}
\usepackage{amsmath}
\usepackage{hyperref}
\usepackage{orcidlink}
\usepackage{balance}

\usepackage{acro}

\DeclareAcronym{dl}{
  short = DL,
  long = deep learning
}
\DeclareAcronym{ekf}{
  short = EKF,
  long = extended Kalman filter
}
\DeclareAcronym{esekf}{
  short = ES-EKF,
  long = error-state extended Kalman filter
}
\DeclareAcronym{enu}{
  short = ENU,
  long = east-north-up
}
\DeclareAcronym{gicp}{
  short = GICP,
  long = generalized iterative closest point
}
\DeclareAcronym{gnss}{
  short = GNSS,
  long = global navigation satellite system
}
\DeclareAcronym{lidar}{
  short = LiDAR,
  long = light detection and ranging
}
\DeclareAcronym{radar}{
  short = RADAR,
  long = radio detection and ranging
}
\DeclareAcronym{icp}{
  short = ICP,
  long = iterative closest point
}
\DeclareAcronym{imu}{
  short = IMU,
  long = inertial measurement unit
}
\DeclareAcronym{ins}{
  short = INS,
  long = inertial navigation system
}
\DeclareAcronym{obd2}{
  short = OBDII,
  long = on-board diagnostics II
}
\DeclareAcronym{obms}{
  short = OBMS,
  long = on-board motion sensors
}
\DeclareAcronym{rgb}{
  short = RGB,
  long = red green blue
}
\DeclareAcronym{rmse}{
  short = RMSE,
  long = root mean square error
}
\DeclareAcronym{slam}{
  short = SLAM,
  long = simultaneous localization and mapping
}
\DeclareAcronym{vslam}{
  short = VSLAM,
  long = visual simultaneous localization and mapping
}
\DeclareAcronym{stor}{
  short = STOR,
  long = semantic and transient object removal
}
\DeclareAcronym{vmr}{
  short = VMR,
  long = visual map registration
}
\DeclareAcronym{vo}{
  short = VO,
  long = visual odometry
}
\DeclareAcronym{maxae}{
  short = MaxAE,
  long = maximum absolute error
}
\DeclareAcronym{sor}{
  short = SOR,
  long = statistical outlier removal
}
\DeclareAcronym{coco}{
  short = COCO,
  long = common objects in context
}
\DeclareAcronym{rcnn}{
  short = R-CNN,
  long = region-based convolutional neural network
}
\DeclareAcronym{navinst}{
  short = NavINST,
  long = navigation and instrumentation research lab
}
\DeclareAcronym{utm}{
  short = UTM,
  long = universal transverse Mercator
}
\DeclareAcronym{vio}{
  short = VIO,
  long = visual-inertial odometry
}
\DeclareAcronym{viwo}{
  short = VIWO,
  long = visual-inertial-wheel odometry
}
\DeclareAcronym{hd}{
  short = HD,
  long = high-definition
}
\DeclareAcronym{adas}{
  short = ADAS,
  long = advanced driver-assistance systems
}
\DeclareAcronym{rem}{
  short = REM,
  long = road experience management
}
\DeclareAcronym{ransac}{
  short = RANSAC,
  long = random sample consensus
}
\DeclareAcronym{roi}{
  short = ROI,
  long = region of interest
}
\DeclareAcronym{ppp}{
  short = PPP,
  long = precise point positioning
}
\DeclareAcronym{rtk}{
  short = RTK,
  long = real time kinematics
}
\DeclareAcronym{mems}{
  short = MEMS,
  long = micro-electro-mechanical systems
}
\DeclareAcronym{ros}{
  short = ROS,
  long  = robot operating system}

  \DeclareAcronym{ppk}{
  short = PPK,
  long = post-processed kinematic
}

\begin{document}

\title{Look to Locate: Vision-Based Multisensory Navigation with 3-D Digital Maps for GNSS-Challenged Environments}

\author{
        Ola Elmaghraby\orcidlink{0009-0007-9528-0488}, 
        Eslam Mounier\orcidlink{0000-0002-7020-2842},~\IEEEmembership{Graduate Student Member,~IEEE,}
        Paulo Ricardo Marques de Araujo\orcidlink{0000-0002-8027-5578},~\IEEEmembership{Member,~IEEE,}
        and~Aboelmagd~Noureldin\orcidlink{0000-0001-6614-7783},~\IEEEmembership{Senior~Member,~IEEE}
        \thanks{This research is supported by grants from the Natural Sciences and Engineering Research Council of Canada (NSERC) under grant numbers: RGPIN-2020-03900 and ALLRP-560898-20. \textit{(Corresponding author: Ola~Elmaghraby.)}}
        \thanks{Ola Elmaghraby and Paulo de Araujo are with the Department of Electrical and Computer Engineering, Queen’s University, Kingston, ON K7L 3N6, Canada (e-mail: ola.elmaghraby.a@queensu.ca; paulo.araujo@queensu.ca).}%
        \thanks{Eslam Mounier is with the Department of Electrical and Computer Engineering, Queen’s University, Kingston, ON K7L 3N6, Canada, and also with the Department of Computer and Systems Engineering, Ain Shams University, Cairo 11535, Egypt (e-mail:eslam.abdelmoneem@queensu.ca).}
        \thanks{Aboelmagd Noureldin is with the Department of Electrical and Computer Engineering, Royal Military College of Canada, Kingston, ON K7K 7B4, Canada, and also with the School of Computing, Queen's University, Kingston, ON K7L 2N8, Canada,  (e-mail:aboelmagd.noureldin@rmc.ca).}
}

\maketitle

\begin{abstract}
In Global Navigation Satellite System (GNSS)-denied environments such as indoor parking structures or dense urban canyons, achieving accurate and robust vehicle positioning remains a significant challenge. This paper proposes a cost-effective, vision-based multi-sensor navigation system that integrates monocular depth estimation, semantic filtering, and visual map registration (VMR) with 3-D digital maps.
Extensive testing in real-world indoor and outdoor driving scenarios demonstrates the effectiveness of the proposed system, achieving sub-meter accuracy 92\% indoors and more than 80\% outdoors, with consistent horizontal positioning and heading average root mean-square errors of approximately 0.98~m and  1.25$^\circ$, respectively.
Compared to the baselines examined, the proposed solution significantly reduced drift and improved robustness under various conditions, achieving positioning accuracy improvements of approximately 88\% on average.
This work highlights the potential of cost-effective monocular vision systems combined with 3D maps for scalable, GNSS-independent navigation in land vehicles.
\end{abstract}


\section{Introduction}
\IEEEPARstart{P}{ositioning} is a cornerstone of autonomous driving, enabling vehicles to plan, control, and make decisions \cite{High_Precision_ositioning_mining}. While \ac{gnss} technologies provide high accuracy positioning capabilities in open-sky environments~\cite{gnss_rtk}, they become unreliable or even denied in environments such as dense urban areas, tunnels, and underground parking \cite{gnss_limitation}. To compensate for \ac{gnss} limitations, some approaches employ high-resolution \ac{lidar}-based positioning systems~\cite{lidar_slam} or integrate high-grade \ac{ins}~\cite{tactical_ins}. Although these solutions can provide accurate and reliable positioning, their high cost hinders their practicality for consumer-level deployment.

In contrast, cameras offer a cost-effective, lightweight, and widely available sensing modality. They are already integrated in most modern vehicles, such as dashcams and \ac{adas} systems, and provide rich visual information useful for navigation and perception~\cite{ mono_cost_effective}. Among camera-based systems, monocular setups are particularly attractive due to their simplicity and minimal hardware requirements.

 However, they face inherent challenges such as scale ambiguity, accumulated drift, and sensitivity to lighting variations or dynamic scenes~\cite{scale_amb_mono}. Recent advances in \ac{dl}, especially in monocular depth estimation, have enabled the extraction of metric-scale 3-D structure from single images~\cite{hu2024metric3d}. 
 This capability paves the way for the innovative use of monocular cameras in applications such as \ac{slam} and map-based navigation.
 
 In parallel, the increasing availability of \ac{hd} and 3-D digital maps, when combined with perception sensors, opens new avenues for achieving high-accuracy map-based positioning \cite{map_localization}. Specifically, by aligning camera-derived 3-D information to pre-existing maps, a globally consistent navigation solution can be achieved. These \ac{dl}-based developments establish monocular solutions as a promising cost-effective alternative to robust positioning, particularly in \ac{gnss}-challenged environments.

Building on these observations, this work addresses the challenge of achieving cost-effective, accurate, and drift-free positioning in GNSS-challenged environments through the development of a vision-based map alignment positioning solution. The paper specifically focuses on the integration of monocular depth estimation with pre-existing 3-D maps to obtain absolute pose corrections integrated within an \ac{esekf} framework.

The contributions of this work are as follows:
\begin{enumerate}
    \item Introducing a \ac{dl}-based monocular 3-D point cloud generation pipeline, integrating a state-of-the-art depth estimation model enhanced by depth preprocessing and postprocessing, \ac{stor}, and point cloud refinement.
    
    \item Proposing a novel \ac{vmr} pipeline that uniquely registers the generated monocular point cloud to the 3-D digital maps, initialized using prior pose information, to provide accurate pose corrections.

    \item Developing an \ac{esekf}-based fusion framework that integrates the \ac{obms} data with VMR output to produce a smooth positioning solution with reduced drift.
    
    \item Validating the effectiveness of the proposed multi-sensor system through real-world driving tests conducted in both urban and indoor environments.
\end{enumerate}

The remainder of this paper is organized as follows. Section~\ref{sec:related_work} reviews related work. Section~\ref{sec:overview} provides an overview of the proposed system. Section~\ref{sec:pc_gen} describes the monocular 3-D point cloud generation pipeline, while Section~\ref{sec:vmr} introduces the vision-to-map registration pipeline for positioning. Section~\ref{sec:fusion} presents the multi-sensor fusion strategy. Section~\ref{sec:setup} outlines the experimental setup, and Section~\ref{sec:results} reports the results. Finally, Section~\ref{sec:conclusion} concludes the paper.

\section{Related Work}
\label{sec:related_work}

\subsection{Vision-Based Dead Reckoning and Sensor Fusion}
Vision-based navigation systems leverage camera inputs to estimate a vehicle’s pose (position and orientation) and understand its environment~\cite{iot_vision_no_gps}. Two foundational approaches in this domain are \ac{vo}~\cite{vo_ola} and \ac{slam}~\cite{survey_slam}. In the context of this work, the term \ac{slam} is used to specifically refer to \ac{vslam} for simplicity. Both methods operate by detecting, describing, and matching visual features across image sequences to estimate motion and structure. However, these methods suffer from drift, scale ambiguity (particularly for monocular cameras), and, in the case of \ac{slam}, strong reliance on loop closure~\cite{vo_slam_limitations}

To address such challenges, many recent works fuse monocular vision with complementary sensors like \ac{imu}, wheel odometry from \ac{obd2} data, and \ac{lidar}. Optimization-based \ac{vio} systems such as Visual-Inertial Systems (VINS)~\cite{qin2018vinsmono} have proven effective on drones. ~\cite{viwo} developed a \ac{viwo} model and a novel data-driven \ac{dl} method to model and mitigate slippage-induced errors effectively. VA-LOAM~\cite{jung2020valoam} uses \ac{lidar} as the primary sensor for \ac{slam}, enhancing visual feature matching. Other works use factor-graph optimization to integrate cameras, \ac{lidar}, and \ac{imu}~\cite{vo_Camera_Lidar_IMU}, 
or semantic-enhanced SLAM~\cite{lou2023slam} for robust positioning under varying conditions.

Although \ac{vo} and SLAM systems enhanced with sensor fusion significantly improve motion estimation accuracy and robustness, they still primarily provide relative pose estimates unless anchored to known global maps or fused with absolute sensors. Additionally, a monocular camera cannot directly measure absolute scale or depth, and visual methods can falter under poor lighting or repetitive textures~\cite{scale_amb_mono}. This has led to numerous works aimed at improving monocular visual navigation via novel depth estimation techniques and map-based positioning.

\subsection{Monocular Depth Estimation for Metric Scale}
\ac{dl} has enabled significant advances in estimating depth from a single image. Early lightweight models such as FastDepth~\cite{fastdepth} achieved real-time inference on embedded devices. Monodepth2~\cite{monodepth2} introduced self-supervised training but was limited to relative depth due to scale ambiguity.

To address this, geometric constraints were incorporated. MonoPP~\cite{Elazab_2025_WACV} leveraged planar-parallax geometry and fixed camera setups to recover metric scale.
Foundation models such as Depth Anything~\cite{ yang2024depthanythingv2} and ZoeDepth~\cite{zoedepth} combined large-scale pretraining with fine-tuning for robust, generalizable depth estimation.

Metric3D~\cite{yin2023metric3d, hu2024metric3d} addressed scale variation and dataset bias by introducing a canonical camera transform and training on over 8 million images, achieving accurate metric predictions even under mixed-camera conditions.

Due to these advances, the performance gap between stereo and monocular systems has significantly narrowed. While stereo provides triangulated depth, it requires precise calibration, synchronization, and a suitable baseline—which can be limiting in embedded or automotive applications. In contrast, monocular systems are lighter, cheaper, and easier to deploy.

\subsection{Map-Based positioning}
There is a growing industrial shift toward map-based positioning in automotive applications~\cite{kang2020lane}. \ac{hd} maps and semantic vector maps are now used in many modern \ac{adas} and autonomous vehicle stacks. These maps enable absolute positioning by aligning live sensor data—typically camera or \ac{lidar}—with prior environmental representations~\cite{Chalvatzaras2023-mapSurvey, Darweesh2022-HDmapTutorial}.

Indoor environments (e.g., warehouses, parking garages) lack \ac{gnss} entirely. Solutions like Lopes \textit{et al.}~\cite{lopes2024monodepth} align vision-derived free-space maps with 2-D floorplans, while \ac{radar}-based systems~\cite{dawson2023integrated} operate in visually degraded settings.

In outdoor urban driving, visual cues from stable man-made structures enable map-based positioning. Yu \textit{et al.}~\cite{yu2020lidarcam} and Zhao \textit{et al.}~\cite{Zhao2024vector} demonstrate efficient positioning by matching monocular images to vector maps.

Cross-modal alignment between 2-D vision and 3-D maps remains an active area. Learning-based methods like I2D-Loc~\cite{i2dloc} refine initial pose estimates via image-to-\ac{lidar} alignment, while newer works like FreeReg~\cite{freereg2024} use pretrained diffusion models for modality-agnostic descriptors.

\section{System Overview}\label{sec:overview}

\begin{figure}
    \centering
    \includegraphics[width=\linewidth]{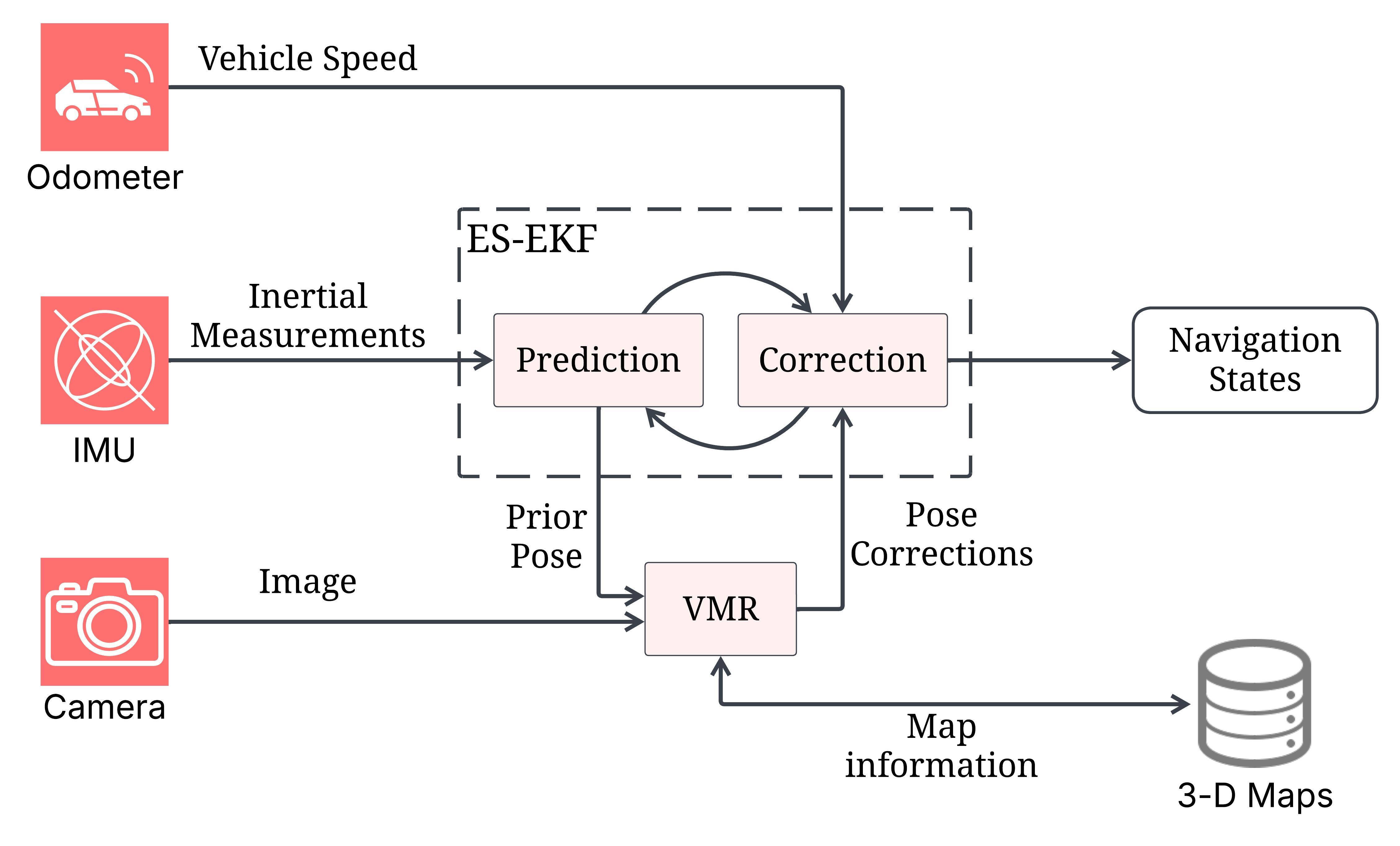}
    \caption{Overview of System Components.}
    \label{fig:system_overview}
\end{figure}

The proposed system integrates monocular vision, inertial sensing, and vehicle odometry within an \ac{esekf} positioning framework to enable resilient and accurate vehicle positioning in \ac{gnss}-challenged environments. For simplicity, we refer to the \ac{esekf} as \ac{ekf} throughout the remainder of this paper. An overview of the system architecture is illustrated in \figurename~\ref{fig:system_overview}.

Monocular images are processed using a \ac{dl}-based pipeline to generate dense 3-D point clouds with metric depth. These point clouds are refined and then aligned with a 3-D reference map through \ac{vmr}, which employs the \ac{gicp} algorithm to compute absolute pose corrections, helping to dimish long-term drift. Simultaneously, inertial measurements from the \ac{imu} are propagated through an \ac{ins} algorithm within the prediction step of the \ac{ekf}. Velocity measurements from the \ac{obd2} interface are used to constrain velocity drifts. The \ac{ekf} continuously integrates these multi-modal measurements to maintain robust estimates of the navigation states. The latest \ac{ekf} estimates are fed back into the \ac{vmr} pipeline to guide subsequent registration steps and into the \ac{ins} algorithm to reset state errors, forming a closed-loop system. 

\section{Deep Learning-Based Monocular 3-D Point Cloud Generation Pipeline} \label{sec:pc_gen}
\begin{figure}
    \centering
    \includegraphics[width=\linewidth]{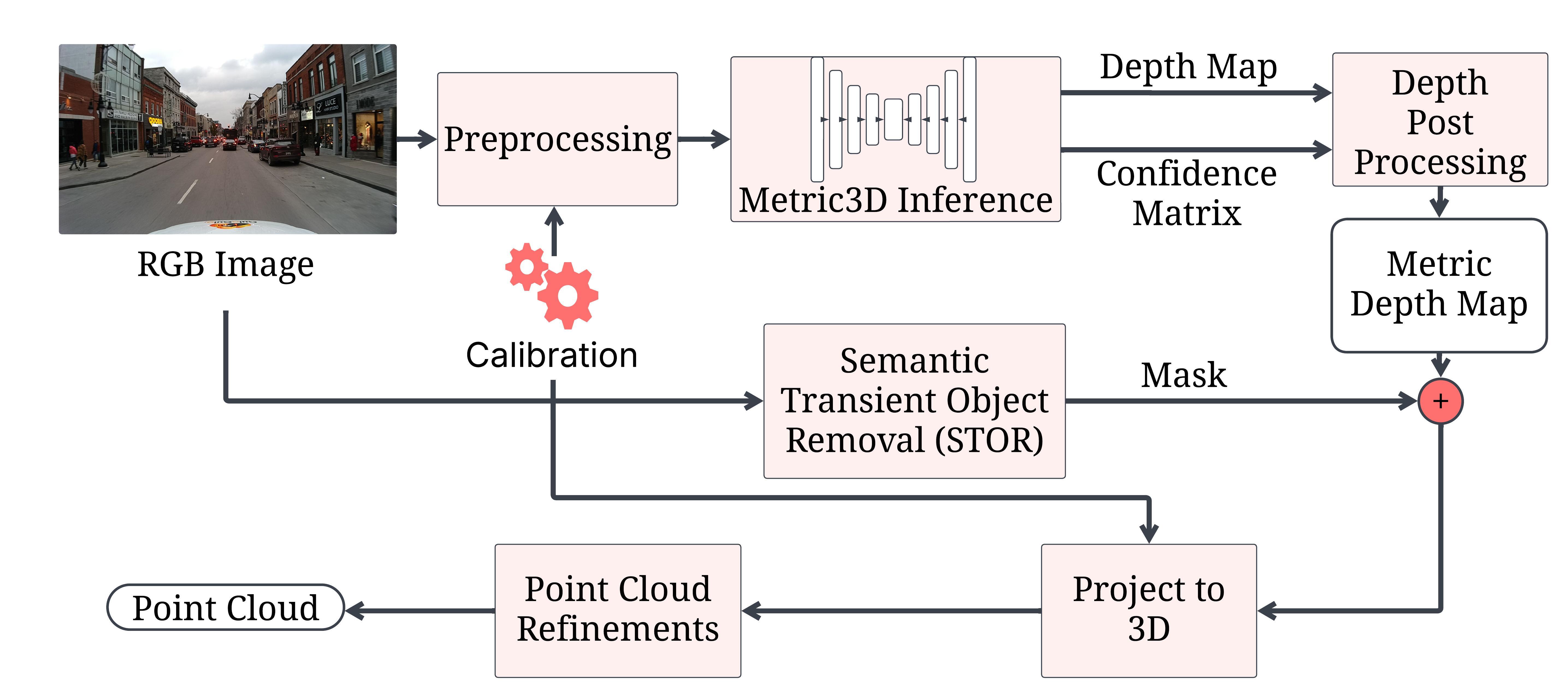}
    \caption{Overview of the point cloud generation process, from depth estimation to filtering.}
    \label{fig:point_cloud_generation}
\end{figure}

This work presents a structured pipeline to generate 3-D point clouds from monocular images, as presented in \figurename~\ref{fig:point_cloud_generation}. The pipeline begins with the depth estimation using a single image and a \ac{dl}-based model. The obtained depth map is then post-processed to remove dynamic objects and correct structural inconsistencies. Finally, the processed depth map is projected onto a 3-D space to generate a point cloud, which is further refined to enhance its quality and reduce computational complexity. All the steps are detailed in the following sections.

\subsection{Preprocessing}
The input \ac{rgb} image is first undistorted using the camera's intrinsic matrix $\mathbf{K}$ and distortion coefficients. The rectified image is resized to a fixed resolution of $616 \times 1064$, maintaining the aspect ratio through symmetric padding with the mean values of the dataset, $\mu$. Channel-wise normalization is then applied using the mean and standard deviation $(\mu, \sigma)$, following standard image preprocessing practices for pretrained \ac{dl} models \cite{img_normalization}.
These preprocessing steps ensure scale-invariant inputs suitable for Metric3Dv2's zero-shot metric depth and surface normal estimation.

\subsection{Depth Estimation Using Metric3D}
The second step in the point cloud generation pipeline involves estimating depth from monocular images. Depth estimation is performed using the Metric3D model \cite{yin2023metric3d,hu2024metric3d}, a monocular depth estimation framework that uses \ac{dl} to infer metric-scale depth maps from single \ac{rgb} images.
To address the challenge of metric depth ambiguity caused by varying camera parameters, particularly focal length, all training data were transformed into a canonical camera space, where images are treated as if captured by a standard camera. This is done either by rescaling depth labels or by resizing input images, effectively normalizing focal length differences. This approach allows the model to learn consistent metric depth across diverse cameras without embedding camera intrinsics in the network.

Metric3D has shown strong performance on standard benchmarks such as KITTI and  datasets, outperforming many existing depth estimation models \cite{hu2024metric3d}. The effectiveness of the model was demonstrated in terms of addressing scale ambiguity and producing accurate metric depth from monocular images. However, additional processing steps are required to refine the depth information and improve geometric consistency for downstream applications.

\subsection{Depth Post Processing}
Metric3D produces two primary outputs: the predicted depth map and a corresponding confidence matrix that assigns a confidence score to each pixel ranging from $0$ to $1$. Low-confidence regions were observed in areas corresponding to the sky and distant in small edges. To mitigate the influence of unreliable predictions, a confidence threshold of \(0.75\) was applied, removing depth values associated with confidence scores below this threshold. 

Additionally, since Metric3D generates predictions in a canonical space, a transformation is required to map the predicted depth back to the camera coordinate system. This transformation is performed using the following scaling relation:
\begin{equation}
\textbf{D}_{\text{metric}} = \textbf{D}_{\text{pred}} \times \frac{f_{\text{input}}}{f_{\text{canon}}}, 
\end{equation}
\noindent where $\textbf{D}_{\text{pred}}$ is the raw depth prediction from Metric3D, $f_{\text{canon}}$ is the training canonical focal length and $f_{\text{input}}$ is the focal length of the input image. The resulting \(\mathbf{D}_{\text{metric}}\) represents the corrected, camera-consistent metric depth map used in subsequent stages.

\subsection{Semantic Transient Object Removal (STOR)}
Dynamic and transient objects, such as vehicles, pedestrians, and buses, introduce inconsistencies between the generated point cloud and the static reference map \cite{remove_dynamic_obj}, particularly in indoor environments where such objects are absent or appear in different locations. These mismatches can cause significant misalignment during the registration process.
To mitigate this, we implement a semantic filtering step that removes such objects from the depth map prior to point cloud generation.

This method leverages a pre-trained \ac{dl} instance segmentation model, specifically Mask \ac{rcnn} from the Detectron2 framework \cite{wu2019detectron2}. This model was chosen for its high accuracy and robustness in detecting and providing pixel-level segmentation masks for objects in complex urban environments, as a result of training on large-scale datasets such as \ac{coco}.

\begin{figure}
\centering
\subfloat[]{%
 \includegraphics[width= 0.49 \linewidth]{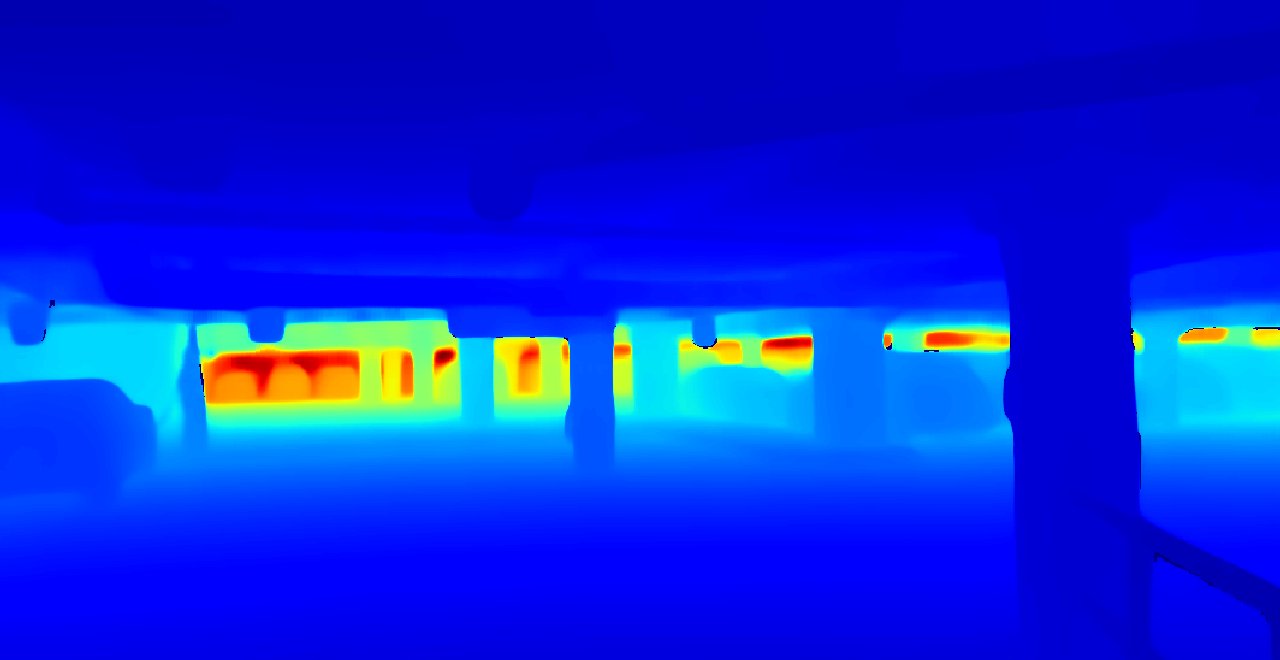}
 \label{fig:with-cars}
} 
\subfloat[]{%
\includegraphics[width= 0.49 \linewidth]{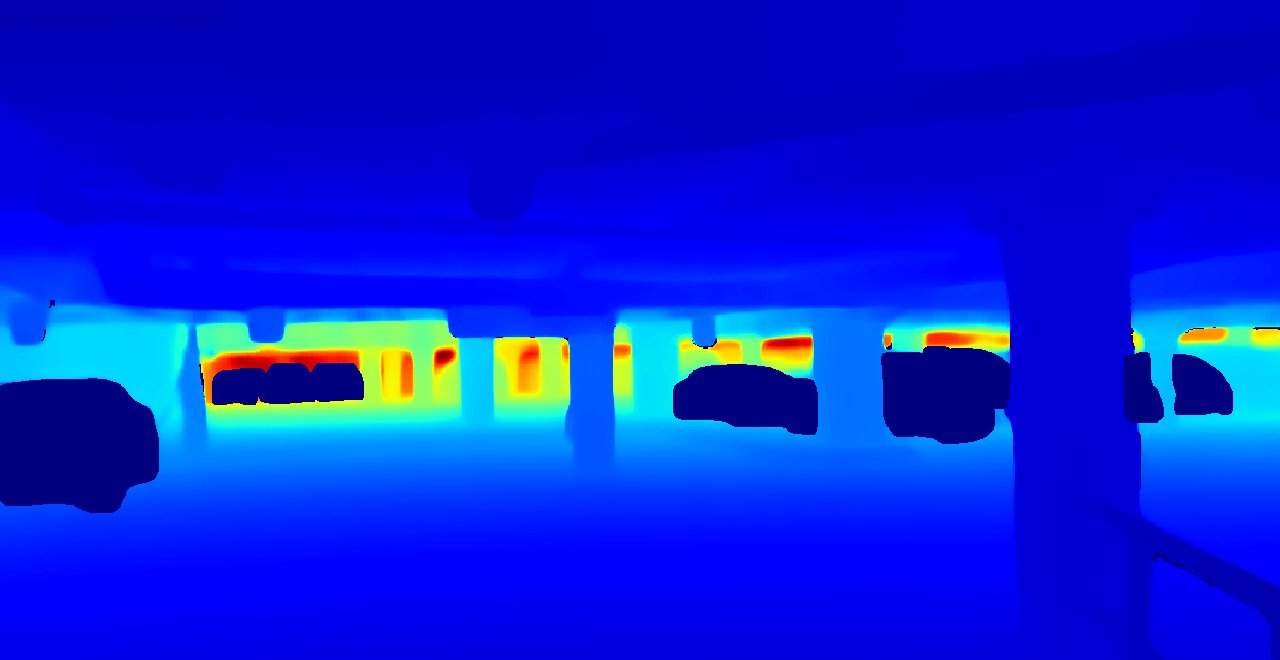}
  \label{fig:witout-cars}
}
\caption{ Demonstration of the effect of the STOR stage. (a) Before filtering. (b) After filtering, the transient objects are dark, indicating 0 depth.}
\label{fig:STOR-demo}
\end{figure}

The model is applied to the \ac{rgb} image to detect and segment different objects, including pedestrians, bicycles, cars, motorcycles, buses, trains, and trucks. After the segmentation masks have been obtained, a dilation filter is applied to the masks to account for potential inaccuracies at the boundaries of the detected objects. The dilation operation expands the mask regions by a predefined kernel size of \((7,7)\), ensuring that any distortions or artifacts around the edges of the objects are included in the mask. The refined segmentation masks are then applied to the depth map to remove the regions corresponding to the objects as presented in \figurename~\ref{fig:STOR-demo}. This is achieved by setting the depth values of the masked regions to zero, effectively excluding these areas from the point cloud generation process.

\subsection{Project to 3-D}
The next step is to generate the point cloud from the processed depth map by: projecting depth values into the 3-D camera frame, aligning the result with the body frame, and applying downsampling to reduce complexity.

\subsubsection{Projection from 2-D Image Plane to 3-D Camera Frame}
Each pixel \((u, v)\) of the depth map, with its corresponding corrected depth \(d\), is projected into the 3-D camera frame using the pinhole camera model, resulting in a 3-D point cloud as illustrated in \figurename~\ref{fig:pc_no_cars}. The intrinsic parameters of the camera, including the focal lengths \(f_x\) and \(f_y\) and the principal point \((c_x, c_y)\), are used to compute the 3-D coordinates in the camera frame, \((X_c, Y_c, Z_c)\), as follows:
\begin{equation}
X_c = \frac{(u - c_x) \cdot d}{f_x}, \quad
Y_c = \frac{(v - c_y) \cdot d}{f_y}, \quad
Z_c = d.
\end{equation}

\subsubsection{Transformation to the Body Frame}
The 3-D points in the camera frame are transformed to the body frame using the extrinsic calibration parameters, which describe the relative pose between the camera and the body frame. This rigid transformation is defined by a rotation matrix \(\mathbf{R}_{\text{calib}} \in SO(3)\) and a translation vector \(\mathbf{t}_{\text{calib}} \in \mathbb{R}^{3}\) that are obtained from the calibration process. The 3-D coordinates in the body frame, \((X_b, Y_b, Z_b)\), are computed as:
\begin{equation}  
\begin{bmatrix}
X_b \\
Y_b \\
Z_b
\end{bmatrix}
= \mathbf{R_{calib}} 
\begin{bmatrix}
X_c \\
Y_c \\
Z_c
\end{bmatrix}
+ \mathbf{t_{calib}} ,
\end{equation}

\subsubsection{Voxel Grid Downsampling}
To reduce computational complexity and storage requirements, a voxel grid downsampling filter is applied to the transformed point cloud. The space was divided into voxels of size $0.2$~m. Each voxel aggregates points within its bounds and approximates them with their centroid. This process preserves the overall structure of the point cloud while significantly reducing the number of points.

\subsection{Point Cloud Refinements}
Despite significant advances in monocular depth estimation \ac{dl}-based models, the resulting point clouds often contain geometric inconsistencies due to prediction uncertainty at long distances, scene artifacts and depth noise around fine structures. To improve registration accuracy and computational efficiency, the point cloud undergoes a three-stage refinement process.

\subsubsection{Point Cloud Cropping}\label{point_cloud_cropping}
Monocular depth estimation is known as an ill-posed task \cite{mono_ill_posed}, where objects in a 2-D image often lack sufficient information to accurately estimate depth. This limitation becomes more pronounced as the distance from the object increases, making it challenging to infer a precise metric depth~\cite{researchgate_depth_error}.

\begin{figure}
\centering
\subfloat[]{%
 \includegraphics[width= 0.49 \linewidth]{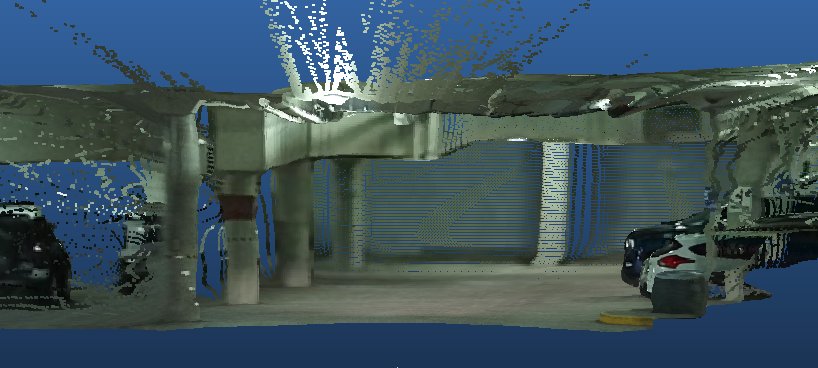}
 \label{fig:pc_original}
} 
\subfloat[]{%
\includegraphics[width= 0.49 \linewidth]{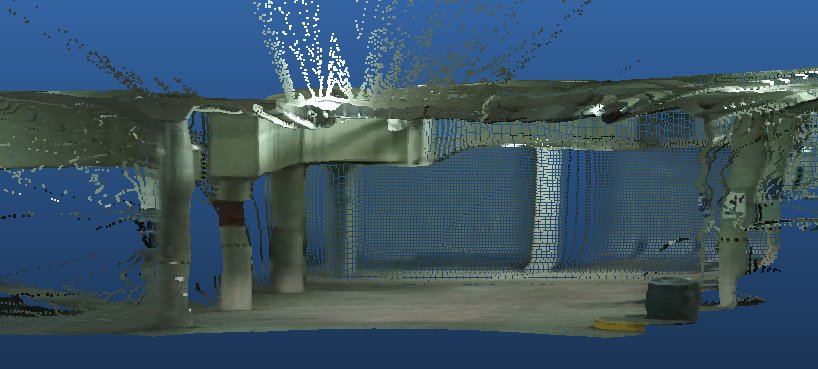}
  \label{fig:pc_no_cars}
}
\\
\subfloat[]{%
\includegraphics[width= 0.49 \linewidth]{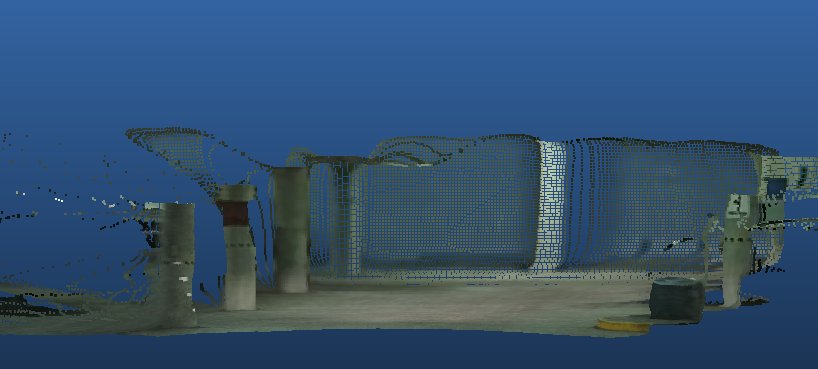}
  \label{fig:pc_crop}
}
\subfloat[]{%
\includegraphics[width= 0.49 \linewidth]{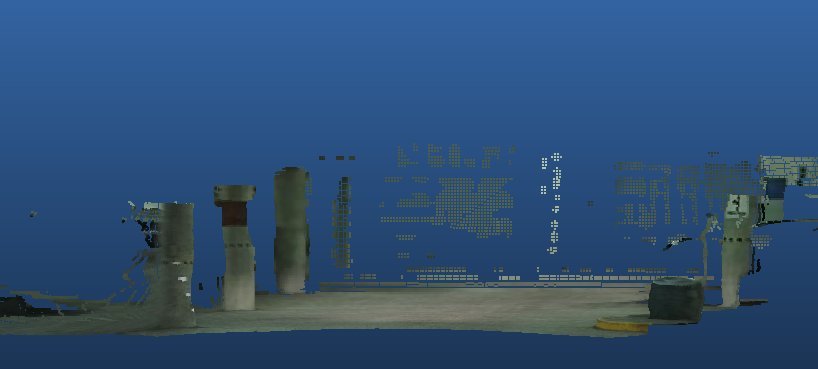}
  \label{fig:pc_filter}
}
\caption{ Point Cloud pipeline output. (a) Original Point Cloud. (b) After STOR. (c) After cropping. (d) After filtering}
\label{fig:pc_result}
\end{figure}

To mitigate the increasing uncertainty in depth estimates with distance, a depth cropping step is applied to exclude points beyond a specified range. Additionally, in indoor environments, ceiling regions often exhibit geometric deformations, e.g., pipes and cable trays, that compromise accuracy. Therefore, points above a height threshold are also removed to suppress ceiling artifacts. The filtered point cloud \( \mathcal{P}_{\text{crop}} \) is obtained by:
\begin{equation}
\mathcal{P}_{\text{crop}} = \left\{ \mathbf{p} \in \mathcal{P} \mid d(\mathbf{p}) < d_{\max},\; h(\mathbf{p}) < h_{\max} \right\},
\end{equation}
\noindent where \( d(\mathbf{p}) \) is the depth of the point \( \mathbf{p} \), \( h(\mathbf{p}) \) is its height, and \( d_{\max} \), \( h_{\max} \) are the predefined depth and height thresholds, respectively. An output example is illustrated in \figurename~\ref{fig:pc_crop}.
\\

\subsubsection{Scale Correction}
Although the point cloud is on the metric scale, a slight scale inconsistency was observed. Empirical evaluation revealed a consistent scale factor, which was quantified by aligning a subset of generated point clouds with their corresponding scans from \ac{lidar} using the \ac{icp} algorithm~\cite{icp_besl1992method}.
The \ac{lidar} scans serve as ground truth references because of their high accuracy and metric consistency. The estimated scale factor, denoted \( s \in \mathbb{R}^{+}\), was then uniformly applied to all point clouds on all trajectories tested. The scaled point cloud \( \mathcal{P}_{\text{scale}} \) is computed as:
\begin{equation}
\mathcal{P}_{\text{scale}} = \left\{ \mathbf{p}' = s \cdot \mathbf{p} \mid \mathbf{p} \in \mathcal{P}_{\text{crop}} \right\},
\end{equation}

\subsubsection{Statistical Outlier Removal}
Although the depth prediction model produces consistent results, it interpolates depth values across region boundaries, reducing edge sharpness, particularly for thin structures such as traffic lights, street poles, and parking garage pillars. When projected into 3-D space, this interpolation introduces distorted ``tails" or elongated artifacts that degrade point cloud quality.

To mitigate these issues, as shown in \figurename~\ref{fig:pc_filter}, a \ac{sor} algorithm is applied. 
For every point \(\mathbf{p}_i\), the mean distance \(\mu_i\) from its \(k\) nearest neighbors is computed. Points that deviate significantly from the global distribution are removed as:
\begin{equation}
\mathcal{P}' = \left\{ \mathbf{p}_i \in \mathcal{P}_{\text{scale}} \mid \mu_i \leq \mu + \tau \cdot \sigma \right\},
\end{equation}
\noindent where \( \mathcal{P}' \) is the output filtered cloud, \(\mu\) and \(\sigma\) are the global mean and standard deviation of the \(\mu_i\) values across all points.

In practice, most of the noise is removed using an initial \ac{sor} pass with conservative parameters (\( k = 6 \), \( \tau = 1.0 \)). However, residual artifacts, particularly in regions with strong depth discontinuities, require a second filtering step with more relaxed settings (\( k = 10 \), \( \tau = 2.0 \)) to effectively eliminate subtle outliers.

\section{Visual Map Registration (VMR)}\label{sec:vmr}
Point cloud registration is used to align the generated point cloud with a reference map to compute absolute pose corrections. These corrections are then fused within the \ac{ekf} framework to mitigate drift and improve long-term accuracy. Before registration, the final output of the point cloud generation step \( \mathcal{P}' \) in the body frame is transformed into the local-level frame using the a priori pose estimate from the \ac{ekf} as:
\begin{equation}
\tilde{\mathcal{P}} = \left\{  \mathbf{T}^{-} \mathbf{p}' \mid \mathbf{p}' \in \mathcal{P}'\right\},
\label{eq:transofrm-pc}
\end{equation}
\noindent where
\[
 \mathbf{T}^{-} =
\begin{bmatrix}
\mathbf{R}^{-} & \mathbf{t}^{-} \\
\mathbf{0} & 1
\end{bmatrix},
\]

\noindent where \( \mathbf{T}^{-} \in SE(3) \) is the transformation matrix from the \ac{ekf} a priori pose estimate, and \( \tilde{\mathcal{P}} \) is the point cloud in the local-level frame.

\begin{figure*}
    \centering
    \includegraphics[width=0.99\linewidth]{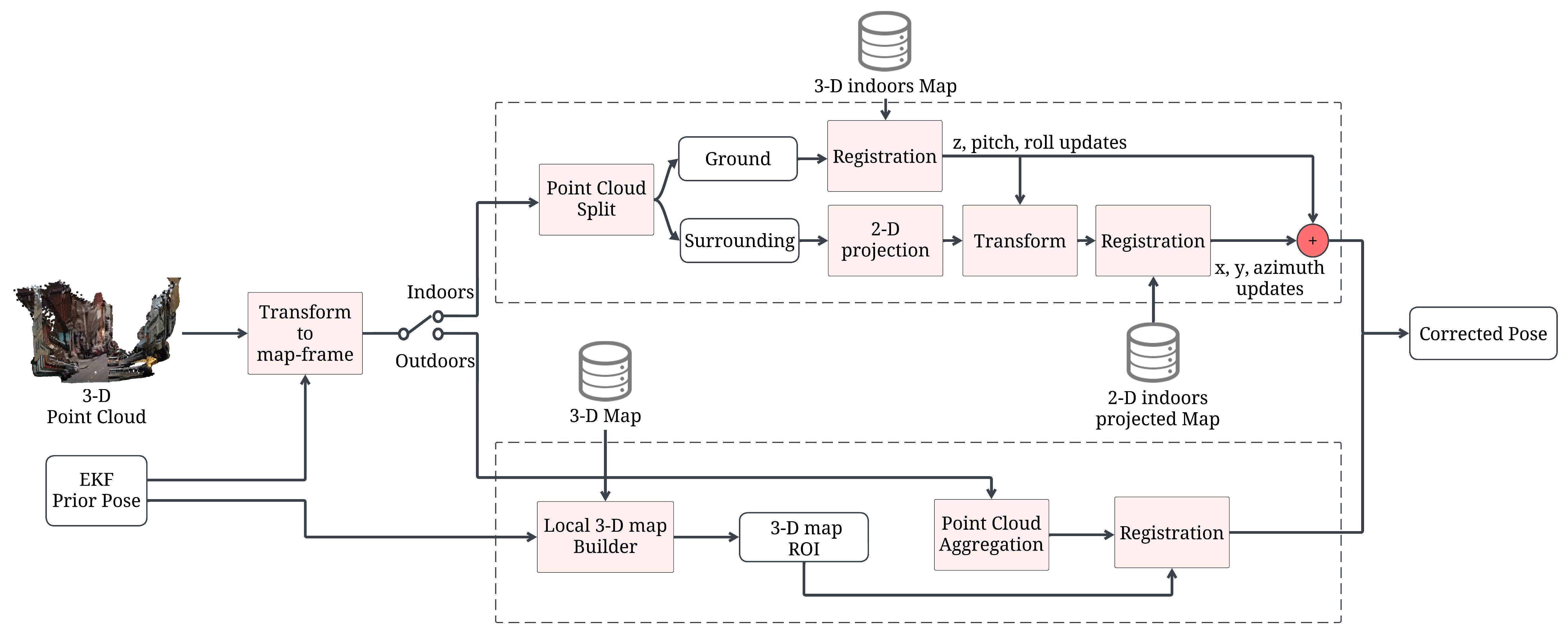}
    \caption{Overview of the indoor and outdoor point cloud registration process.}
    \label{fig:registration_process}
\end{figure*}

The registration process differs between indoor and outdoor environments due to variations in structure and feature distribution. However, both rely on the \ac{gicp} algorithm as the core of the alignment process. 
An overview of the registration workflow for both environments is illustrated in \figurename~\ref{fig:registration_process}.

\subsection{\ac{gicp} Algorithm}

The registration was performed using the \ac{gicp} algorithm, which extends the standard \ac{icp} algorithm by modeling local surface geometry and incorporating uncertainty in the point cloud data through per-point covariance estimates \cite{Segal2009GeneralizedICP}. This extension makes it particularly well-suited for aligning point clouds with varying densities and structures and offers more robustness to noise and outliers compared to the point-to-point and point-to-plane \ac{icp} variants.

\ac{gicp} estimates a rigid transformation \(\Delta\mathbf{T} \in SE(3)\) that aligns the source point cloud, generated from the monocular pipeline, to the target point cloud from the reference 3-D map. 
This transformation is then applied to the EKF prior pose estimate \(\mathbf{T}^{-}\) to obtain the corrected pose$\hat{\mathbf{T}}$ as:
\begin{equation}
    \hat{\mathbf{T}} = \Delta \mathbf{T} \, \mathbf{T}^{-}.
    \label{eq:pose-correction}
\end{equation}

\subsection{Indoor Registration}
Indoor environments, such as parking garages, possess distinct structural features, including floors, walls, pillars, and ceilings, which require additional measures to effectively adapt to and leverage the characteristics of the setting.

\subsubsection{Point Cloud Splitting}
In indoor environments, after the transient objects are removed, the point cloud is dominated by ground points, with fewer points corresponding to the pillars and walls. This imbalance can bias the registration process, causing it to optimize primarily for the ground while neglecting vertical features. To address this, a splitting algorithm was applied to separate the ground points from the surrounding points (e.g., pillars and walls). As illustrated in \tablename~\ref{tab:point_counts}, the average number of ground points in indoor environments far exceeds the average number of surrounding points (after cropping ceiling points) based on a tested trajectory, whereas outdoor environments exhibit a more balanced distribution.

\begin{table}
    \centering
    \caption{Comparison of ground and surrounding point counts for indoor and outdoor environments.}
    \label{tab:point_counts}
    \begin{tabular}{|l|c|c|}
        \hline
        \textbf{Environment} & \textbf{Ground Points} & \textbf{Surrounding Points} \\
        \hline
        Indoor & $180,695$ & $39,599$ \\
        Outdoor &  $324,537$ & $405,677$ \\
        \hline
    \end{tabular}
\end{table}

To further enhance registration quality, a 2-D projection of the surrounding point cloud and the indoor map was applied. This projection was generated from a top-down view, condensing the vertical features (e.g., pillars and walls) into a 2-D plane. The ground points were removed from both the point cloud and the map before projection to focus on the vertical structures. This approach improved the robustness of the registration by emphasizing the dominant vertical features in indoor environments.

\subsubsection{Two Stage Registration}

The registration process is performed in two stages to incrementally refine the alignment of the estimated trajectory. In the first stage, known as ground registration, the ground point cloud is aligned to correct for errors in the vertical \(z\)-axis as well as the pitch ($\theta$) and roll ($\phi$) angles. 

The resulting \ac{gicp} transformation matrix is then decomposed into its Euler angles and translation components. A new transformation matrix $\Delta \mathbf{T}_v$ is reconstructed using only the correction vertical components (\(\Delta z\), \( \Delta \theta\), and \( \Delta \phi\)) and applied to the surrounding point cloud following \eqref{eq:transofrm-pc}.

Next, the vertically aligned surrounding point cloud is registered to the corresponding portion of the map.
This step focuses on refining the components of the horizontal position \(x\) and \(y\), along with the azimuth ($\psi$) angle. The \ac{gicp} output is used to construct a transformation matrix $\Delta \mathbf{T}_h$ using horizontal corrections (\(\Delta x\),\(\Delta y\), and \( \Delta \psi\)) to improve the horizontal alignment without overriding the ground corrections applied earlier.

Finally, the transformations obtained from both registration stages are integrated as:
\begin{equation}
\Delta \mathbf{T} = \Delta \mathbf{T}_h \,  \Delta \mathbf{T}_v,
\end{equation}
\noindent which is then applied to the \ac{ekf} prior pose estimate after \eqref{eq:pose-correction}. The resulting corrected pose $\hat{\mathbf{T}}$ is fused into the \ac{ekf} to improve the overall trajectory estimation.

\subsection{Outdoor Registration} \label{sec:outdoor_registration}
Outdoor trajectories are generally free of spatial constraints, and the corresponding 3-D maps can cover large areas. To efficiently manage this, the outdoor maps in this work are organized into a geospatial database structure, as described in Section~\ref{sec:3D-maps}. For each registration epoch, a local subset of the map is retrieved based on the vehicle's current position. Following the approach in~\cite{eslam_lidar_fusion}, a \ac{roi} map of size \(100 \times 100\) meters is centered on the latest \ac{ekf} position estimate. This \ac{roi} is assembled by querying the offline geospatial database, which stores pre-aligned 3-D \texttt{.las} files containing high-resolution and georeferenced \ac{lidar} scans of the environment. The resulting \ac{roi} map, denoted as $\mathcal{P}_{\text{ref}}$, serves as reference for the alignment process.

Unlike indoor environments, outdoor scenes typically exhibit a more balanced distribution of features across ground and vertical structures (e.g., buildings, trees, signs). Additionally, since the ceiling is cropped in indoor environments, the depth threshold used for cropping in outdoor settings is higher than that in indoor scenarios, affecting the number of points, as illustrated in \tablename~\ref{tab:point_counts}. As a result, the point cloud splitting step is not required. However, due to the generally larger depth ranges encountered outdoors, monocular depth estimation tends to suffer from an increased error at longer distances, as discussed in Section~\ref{point_cloud_cropping}. To mitigate this, we aggregate point clouds across multiple frames based on displacement. This aggregation helps smooth out local depth estimation errors. 

The aggregation process starts with an empty point cloud buffer and monitors vehicle displacement using the \ac{ekf} trajectory. When the displacement from the last aggregation position exceeds a certain threshold 
$d_{\text{min}}$, it is transformed into the world frame using the corresponding \ac{ekf} pose, following \eqref{eq:transofrm-pc}, and its points added to the point cloud buffer. 
Aggregation continues until the total distance traveled exceeds a threshold $d_{\text{max}}$, after which the buffer points are merged, yielding a point cloud denoted as $\mathcal{P}_{\text{merged}}$. Then, voxel grid filtering is applied to downsample the aggregated point cloud.

A full 3-D \ac{gicp} registration is then performed to align the aggregated source point cloud $\mathcal{P}_{\text{merged}}$ with the target reference map $\mathcal{P}_{\text{ref}}$, yielding the correction transformation matrix $\Delta \mathbf{T}$. The updated pose matrix $\hat{\mathbf{T}}$ is obtained following \eqref{eq:pose-correction}, which corrects all six pose states (\(x\), \(y\), \(z\), \(\theta\), \(\phi\), \(\psi\)), and is subsequently incorporated into the \ac{ekf} correction step.

\section{Multi Sensor Fusion}\label{sec:fusion}
The proposed sensor fusion framework integrates measurements from onboard sensors, including an \ac{imu} and vehicle speed via \ac{obd2}, along with corrections from the \acf{vmr} module, to estimate the pose of a land vehicle. While all sensor measurements are acquired in the body frame, state estimation is performed in the local \ac{enu} navigation frame. The fusion relies on an \ac{ekf}, where the nominal state is propagated using \ac{ins} mechanization, and external measurements provide corrections to position, velocity, and orientation.

\subsection{Inertial Navigation Systems (INS) Mechanization}
\label{sec:INS}
The \ac{ins} estimates the position, velocity, and orientation of a moving platform by integrating inertial measurements from the \ac{imu} \cite{araujo2022evaluation}. The system state consists of orientation \( \mathbf{R}_k \in \mathrm{SO}(3) \), velocity \( \mathbf{v}_k \in \mathbb{R}^3 \), and position \( \mathbf{p}_k \in \mathbb{R}^3 \), all expressed in the local \ac{enu} navigation frame.

The measured linear acceleration \( \mathbf{f}_{k-1} \) and the angular velocity \( \boldsymbol{\omega}_{k-1} \) are first corrected by their respective biases \( \mathbf{b}_{f,{k-1}} \) and  \( \mathbf{b}_{\omega,{k-1}} \). 
Orientation is updated using a quaternion-based integration of the corrected angular increment. The resulting orientation \( \mathbf{R}_k \) is used to transform the corrected acceleration from the body frame to the navigation frame. The velocity and position are then updated using first- and second-order integration, respectively:
\begin{align}
\mathbf{R}_k &= \mathbf{R}_{k-1} \otimes \text{quat}\left( \tfrac{1}{2} \left( \boldsymbol{\omega}_{k-1} - \mathbf{b}_{\omega,{k-1}} \right) \Delta t \right), \label{eq:ins_rot_quat} \\
\mathbf{v}_k &= \mathbf{v}_{k-1} + \left( \mathbf{R}_{k-1} (\mathbf{f}_{k-1} - \mathbf{b}_{f,{k-1}}) + \mathbf{g} \right) \Delta t, \label{eq:ins_vel} \\
\mathbf{p}_k &= \mathbf{p}_{k-1} + \mathbf{v}_{k} \Delta t . 
\label{eq:ins_pos}
\end{align}
\noindent where \( \text{quat}(\cdot) \) represents the incremental quaternion constructed from the angular rate vector, \( \otimes \) denotes the quaternion product and \( \mathbf{g} \) is the gravity vector in the \ac{enu} navigation frame.

\subsection{Velocity Corrections Using OBDII Module}
To constrain the velocity drift from the \ac{ins} mechanization, the \ac{obd2} interface provided an independent and reliable source of velocity information, which was incorporated into the \ac{ekf} fusion. Since \ac{obd2} provides a scalar speed (\(v_{\text{odometer}}\)), representing the vehicle's longitudinal motion, this value must be projected into the \ac{enu} navigation frame to be compatible with the \ac{ekf} state vector \cite{eslam_lidar_fusion}.

Using the estimated vehicle's azimuth (\(\psi\)) and pitch (\(\theta\)) angles, the velocity components in the East (\(v_x\)), North (\(v_y\)), and Up (\(v_z\)) are derived as:
\begin{equation}
\begin{aligned}
v_x &= \sin(\psi) \cos(\theta) v_{\text{odometer}}, \\
v_y &= \cos(\psi) \cos(\theta) v_{\text{odometer}}, \\
v_z &= \sin(\theta) v_{\text{odometer}}.
\end{aligned}
\end{equation}

\subsection{Extended Kalman Filter (EKF) Design}
The \ac{ekf} is used to estimate the error state \(\delta \mathbf{x}\), which represents the deviation of the nominal state from the true state.

The error state vector at epoch \(k\) is defined as:
\begin{equation}
\delta \mathbf{x}_k = \begin{bmatrix}
{\delta \mathbf{p}_k}^\top &
{\delta \mathbf{v}_k}^\top &
{\delta \boldsymbol{\theta}_k}^\top &
{\delta \mathbf{b}_{f,k}}^\top &
{\delta \mathbf{b}_{w,k}}^\top 
\end{bmatrix}^\top,
\end{equation}
\noindent where $\delta \mathbf{p}_k$, $\delta \mathbf{v}_k$, and $\delta \boldsymbol{\theta}_k$ denote the position, velocity, and attitude errors, respectively. The terms $\delta \mathbf{b}_{f,k}$ and $\delta \mathbf{b}_{w,k}$ represent the accelerometer and gyroscope bias error.
The nominal state \(\mathbf{x}_k\) is propagated by using the \ac{ins} mechanization as detailed in Section~\ref{sec:INS}.

The prediction step also comprises the computation the a priori covariance \(\mathbf{P}_k^-\) using the state transition matrix \(\mathbf{F}_{x,k-1}\) and process noise covariance \(\mathbf{Q}_{k-1}\):
\begin{equation}
\mathbf{P}_k^- = \mathbf{F}_{x,k-1} \mathbf{P}_{k-1}^+ \mathbf{F}_{x,k-1}^\top + \mathbf{Q}_{k-1},
\end{equation}
\noindent where
{\scriptsize
\begin{align*}
    \mathbf{F}_{x,k-1} &= \begin{bmatrix}
    \mathbf{I}_3 & \mathbf{I}_3 \Delta t & \mathbf{0}_3 & \mathbf{0}_3 & \mathbf{0}_3 \\
    \mathbf{0}_3 & \mathbf{I}_3 & -\mathbf{R}_{k-1}^+ \boldsymbol{\Omega}_{f,  k-1} \Delta t & -\mathbf{R}_{k-1}^+ \Delta t & \mathbf{0}_3 \\
    \mathbf{0}_3 & \mathbf{0}_3 & \exp(-\boldsymbol{\Omega}_{\omega, k-1} \Delta t)^\top & \mathbf{0}_3 & -\mathbf{I}_3 \Delta t \\
    \mathbf{0}_3 & \mathbf{0}_3 & \mathbf{0}_3 & \mathbf{I}_3 & \mathbf{0}_3 \\
    \mathbf{0}_3 & \mathbf{0}_3 & \mathbf{0}_3 & \mathbf{0}_3 & \mathbf{I}_3
    \end{bmatrix}, \\[4pt]
    \boldsymbol{\Omega}_{\omega,  k-1} &= \text{skew}(\boldsymbol{\omega}_{k-1} - \mathbf{b}_{w,k-1}^+), \\[4pt]
    \boldsymbol{\Omega}_{f,  k-1} &= \text{skew}(\mathbf{f}_{k-1} - \mathbf{b}_{f,k-1}^+),
\end{align*}
}
\noindent and $\text{skew}(\cdot) \in \mathbb{R}^{3 \times 3}$ denotes the skew-symmetric matrix generated from a three-dimensional vector.

The process noise covariance \(\mathbf{Q}_{k-1}\) models \ac{imu} sensor noise and biases:
\begin{equation}
\mathbf{Q}_{k-1} = \text{diag}\left(
\begin{matrix}
\sigma_f^2 \Delta t^2 \mathbf{I}_3 & \sigma_w^2 \Delta t^2 \mathbf{I}_3 & \sigma_{b_f}^2 \Delta t \mathbf{I}_3 & \sigma_{b_w}^2 \Delta t \mathbf{I}_3
\end{matrix}
\right).
\end{equation}
\noindent where the values of these noise parameters were estimated using Allan Variance analysis, a widely used technique for \ac{imu} noise characterization \cite{el2004allantools}.

The correction step updates the state using external measurements \(\mathbf{z}_k\). In this system, two sources of corrections are applied independently: the speed correction from the odometer and the pose correction from VMR. The speed correction is applied first to constrain the velocity states, after which the corrected pose is passed to the VMR module. When an \ac{icp}-based registration measurement becomes available, it is used to further refine position and orientation estimates.
For both corrections, the measurement model is:
\begin{equation}
\mathbf{z}_k = \mathbf{H}_k \delta \mathbf{x}_k + \boldsymbol{\eta}_k, \quad \boldsymbol{\eta}_k \sim \mathcal{N}(0, \mathbf{N}_k),
\end{equation}
where \(\mathbf{H}_k = [\mathbf{0}_3 \,\, \mathbf{I}_3 \,\, \mathbf{0}_3 \,\, \mathbf{0}_3 \,\, \mathbf{0}_3]\) for the speed correction, and
$$
\mathbf{H}_k = 
\begin{bmatrix}
    \mathbf{I}_3 & \mathbf{0}_3 & \mathbf{0}_3 & \mathbf{0}_3 & \mathbf{0}_3 \\
    \mathbf{0}_3 & \mathbf{0}_3 & \mathbf{I}_3 & \mathbf{0}_3 &  \mathbf{0}_3
\end{bmatrix},
$$

It should be noted that the measurement covariance, \(\mathbf{N}_k\), was empirically derived from uncertainty analysis of the \ac{obd2}-based vehicle speed measurements and monocular depth estimation using offline test data.

Finally, the Kalman gain \(\mathbf{K}_k\) and update equations are executed as:
\begin{equation}
\mathbf{K}_k = \mathbf{P}_k^- \mathbf{H}_k^\top \left( \mathbf{H}_k \mathbf{P}_k^- \mathbf{H}_k^\top + \mathbf{N}_k \right)^{-1}.
\end{equation}
\begin{equation}
\delta \mathbf{x}_k^+ = \delta \mathbf{x}_k^- + \mathbf{K}_k (\mathbf{z}_k - \mathbf{H}_k \delta \mathbf{x}_k^-),
\end{equation}
\begin{equation}
\mathbf{P}_k^+ = (\mathbf{I} - \mathbf{K}_k \mathbf{H}_k) \mathbf{P}_k^- (\mathbf{I} - \mathbf{K}_k \mathbf{H}_k)^\top + \mathbf{K}_k \mathbf{N}_k \mathbf{K}_k^\top.
\end{equation}

\subsection{Dynamic Tuning}  \label{sec:dynamic_tuning}
To improve robustness against measurement outliers, the measurement noise covariance \(\mathbf{N}_k\) is adaptively tuned based on the magnitude of the correction. The principle is to reduce certainty in measurements that deviate significantly from the current prior estimate.

The measurement covariance matrix \( \mathbf{N} \) is dynamically adjusted using a tuning vector \( \boldsymbol{\lambda} \)
defined as:
\begin{equation}
\boldsymbol{\lambda} = \begin{bmatrix} \lambda_x & \lambda_y & \lambda_z & \lambda_\theta & \lambda_\phi & \lambda_\psi \end{bmatrix}^\top,
\end{equation}
\noindent where \( \lambda_x \), \( \lambda_y \), and \( \lambda_z \) are the tuning factors for the East (\(x\)), North (\(y\)), and Up (\(z\)) directions, respectively, and \( \lambda_\theta \), \( \lambda_\phi \), and \( \lambda_\psi \) represent the tuning factors for pitch, roll, and azimuth, respectively.

The tuning factors for all components are computed as:
\begin{equation}
\lambda_i = \exp\left(\alpha \cdot \left| \Delta i \right|\right), \quad \text{for } i \in \{x, y, z, \theta, \phi, \psi\},
\end{equation}
\noindent where \( \Delta i = i_{\text{new}} - i_{\text{est}} \) denotes 
the correction for each state component. 
The scaling hyperparameter \( \alpha \) was empirically tuned to balance responsiveness with stability in the presence of measurement uncertainty.

The dynamically adjusted measurement covariance matrix \( \mathbf{N}_{\text{dynamic}} \) is then computed as:
\begin{equation}
\mathbf{N}_{\text{dynamic}} = \mathbf{N}_{\text{static}} \cdot \text{diag}(\boldsymbol{\lambda}),
\end{equation}
\noindent where \( \mathbf{N}_{\text{static}} \) is the nominal measurement noise covariance matrix.

\section{Experimental Setup And Maps Dataset}\label{sec:setup}

\subsection{Sensor Configuration and Data Collection}

\begin{figure}
    \centering
\includegraphics[ width=\columnwidth]{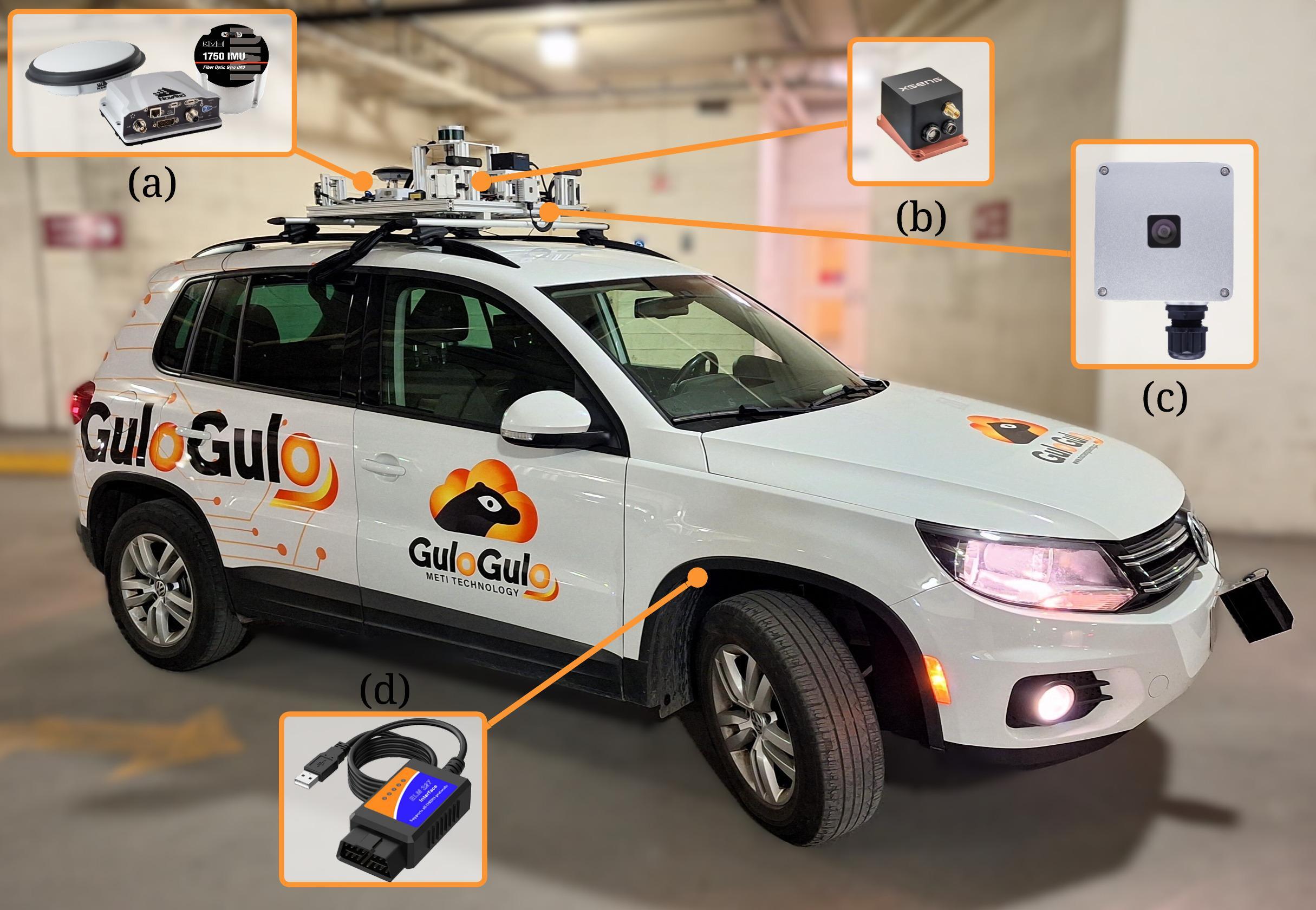}
    \caption{NavINST Multi-Sensor Platform highlighting the primary sensors. (a) NovAtel PwrPak7-E1, (b) Xsens IMU, (c) Luxonis OAK-1 W PoE Camera, and (d) OBDII Module}
    \label{fig:setup}
\end{figure}

The experimental evaluation was conducted using the \ac{navinst} platform~\cite{navinst}, a research vehicle equipped with a tightly synchronized multi-sensor suite to support positioning research in both outdoor and indoor environments. A subset of the available sensors, highlighted in \figurename~\ref{fig:setup}, was the focus of this study. All data were recorded using \ac{ros}, then temporally aligned to a common frequency of 10~Hz to reflect the relatively slow vehicle dynamics in all scenarios tested.

The primary sensors used include:
\begin{itemize}
    \item A Luxonis OAK-1 W PoE monocular \ac{rgb} camera operating at 30~Hz.
    \item An Xsens \ac{mems}-grade \ac{imu} that provides inertial measurements at 100~Hz.
    \item An \ac{obd2} interface providing the vehicle’s forward speed at 16~Hz.
    \item For outdoor experiments, ground truth was obtained at 50~Hz from a NovAtel PwrPak7-E1 system, comprising a high-end \ac{gnss} receiver, tactical-grade \ac{imu} (KVH-1750), with \ac{ppk} corrections.
    \item For indoor experiments, where \ac{gnss} is unavailable, ground truth was obtained at 50~Hz from a multi-sensor system that integrates tactical-grade \ac{imu}, wheel odometry, and \ac{lidar} registration with a pre-built high-accuracy 3-D map \cite{eslam_lidar_fusion}.
\end{itemize}

\subsection{3-D Maps Dataset}
\label{sec:3D-maps}

\begin{figure}
    \centering
\includegraphics[trim={1cm 1cm 1cm 1cm}, clip,width=1\linewidth]{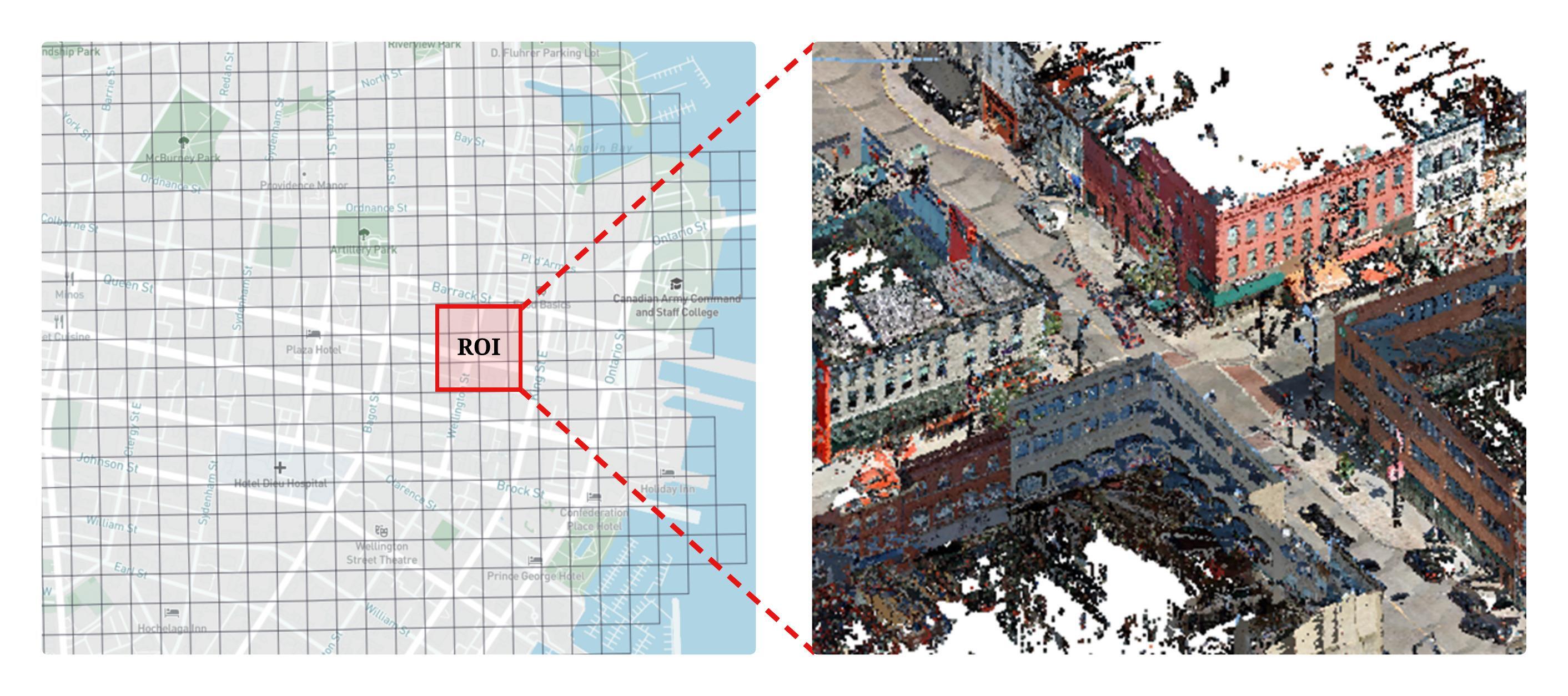}
    \caption{A subset of the city of Kingston 3-D Maps.}
    \label{fig:outdoor_map}
\end{figure}

\begin{figure}
    \centering
\includegraphics[width=0.8\linewidth]{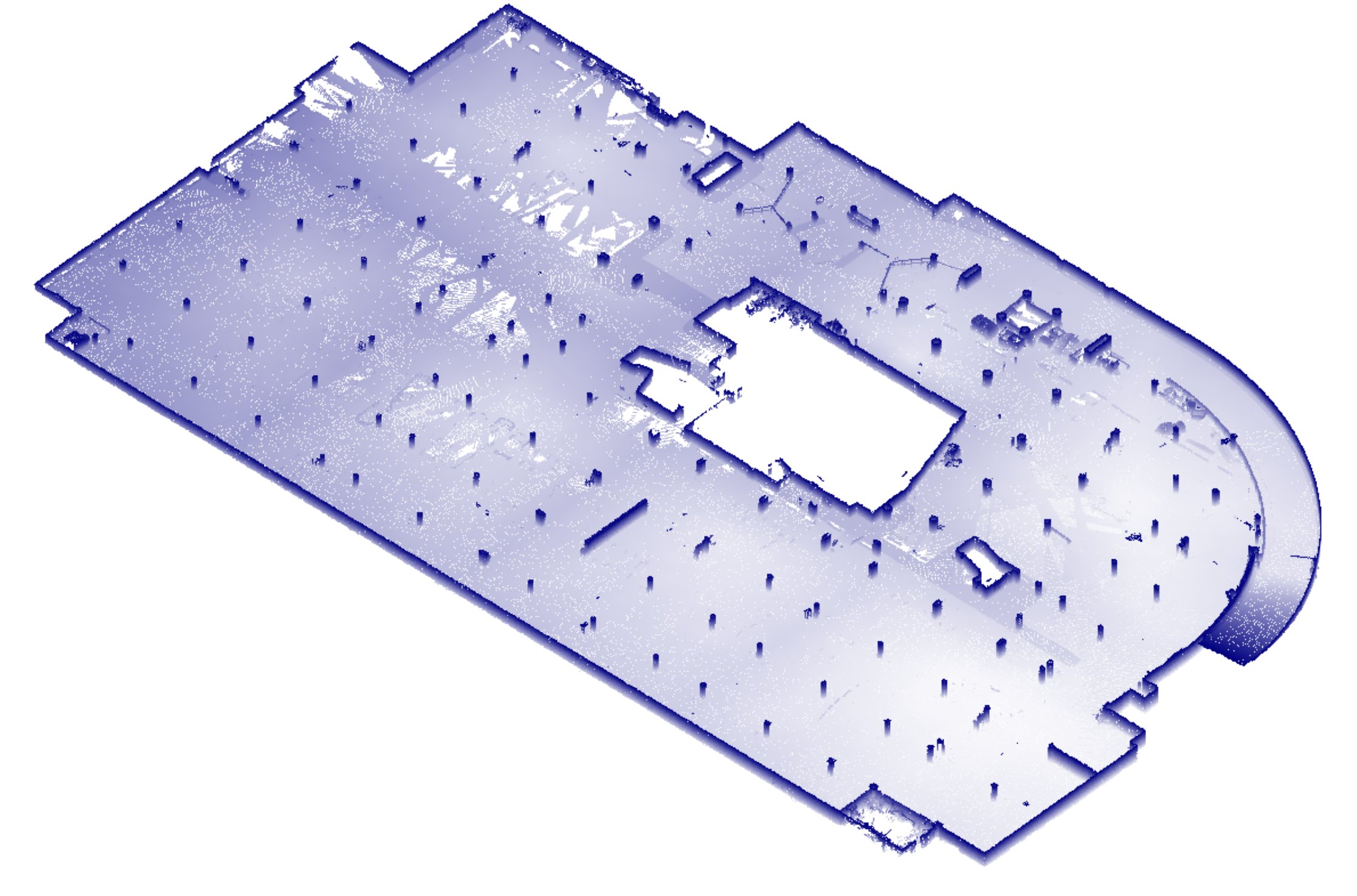}
    \caption{3-D map of the indoor parking garage.}
    \label{fig:indoor_map}
\end{figure}

The outdoor experiments utilize a high-accuracy 3-D map of downtown Kingston, Ontario, Canada. This map was generated through municipal mobile mapping efforts and includes dense street-level point cloud data from urban structures such as buildings, poles, and signage. The data consists of tens of thousands of \texttt{.las} files, each containing millions of 3-D georeferenced points. These 3-D files are organized within a geospatial database structured into spatial tiles with consistent geometry, where each tile references a corresponding 3-D \texttt{.las} file~\cite{mounier2022utilizing}. As shown in \figurename~\ref{fig:outdoor_map}, specifying a spatial \ac{roi} retrieves relevant tiles, which are combined to construct a local 3-D map. These maps serve as prior knowledge for accurate positioning in outdoor driving scenarios.

To validate the proposed method through indoor driving experiments, a 3-D map of an underground parking garage from the NavINST dataset~\cite{navinst}. The 3-D map, shown in \figurename~\ref{fig:indoor_map}, was created using a high-resolution stationary \ac{lidar} scanner and refined through post-processing to remove irrelevant elements, such as stationary vehicles and pedestrians.

\section{Results and Discussion}\label{sec:results}
The proposed method is evaluated in six representative segments: two indoors and four outdoors. These segments were carefully selected to cover a range of conditions, including varying feature density, structural complexity, and dynamic elements such as moving vehicles and pedestrians. For each segment, we compare three positioning approaches. The first is the \ac{obms} \ac{ekf} pipeline, which represents a baseline solution based solely on inertial mechanization corrected by forward speed measurements obtained from the vehicle odometer. This configuration does not incorporate vision-based updates or map registration. 

The second method, referred to as vanilla VMR, implements visual map registration using raw monocular depth and image features, but excludes the enhancements introduced in this work, i.e., STOR, depth cropping, scale correction, point cloud filtering \ref{sec:pc_gen}, distance-based aggregation \ref{sec:outdoor_registration}, and \ac{ekf} dynamic tuning \ref{sec:dynamic_tuning}). 

The third approach is the proposed VMR solution, which integrates all of the aforementioned enhancements to improve robustness and positioning accuracy across diverse environments. This comparison allows us to isolate the contributions and evaluate the overall effectiveness of the full pipeline under real-world constraints.

\subsection{Indoors}
The proposed VMR solution was evaluated on two challenging indoor trajectories recorded within a parking garage. Both segments reflect real-world structured environments characterized by poor lighting, low-texture surfaces, and frequent stop-and-go motions. The first segment (I1), spanning approximately $1.8$ minutes ($110$ seconds), includes several turns, vehicle stops, and slow maneuvers with a peak speed of approximately $15$\,km/h. The second segment (I2) is notably longer, about $4.3$ minutes ($260$ seconds), and includes tight loops and more aggressive maneuvers, with a maximum speed of approximately $18$\,km/h. This increased duration and complexity make segment I2 a more rigorous test of long-term drift resilience.

\begin{figure*}
\centering
\subfloat[]{%
 \includegraphics[height= 0.4 \linewidth]{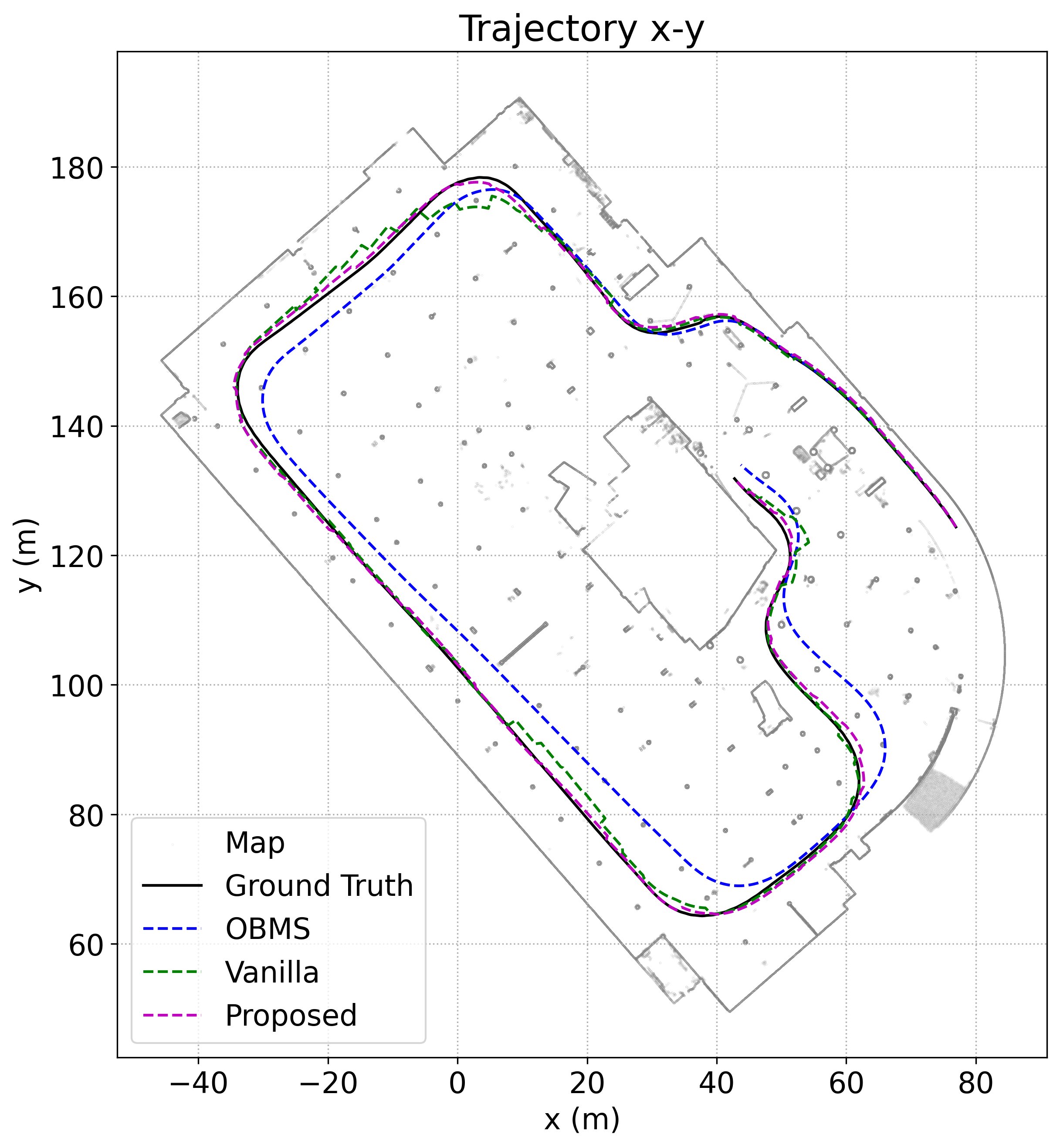}
 \label{fig:segment1_traj}
} 
\hspace{25pt}
\subfloat[]{%
\includegraphics[height= 0.4 \linewidth]{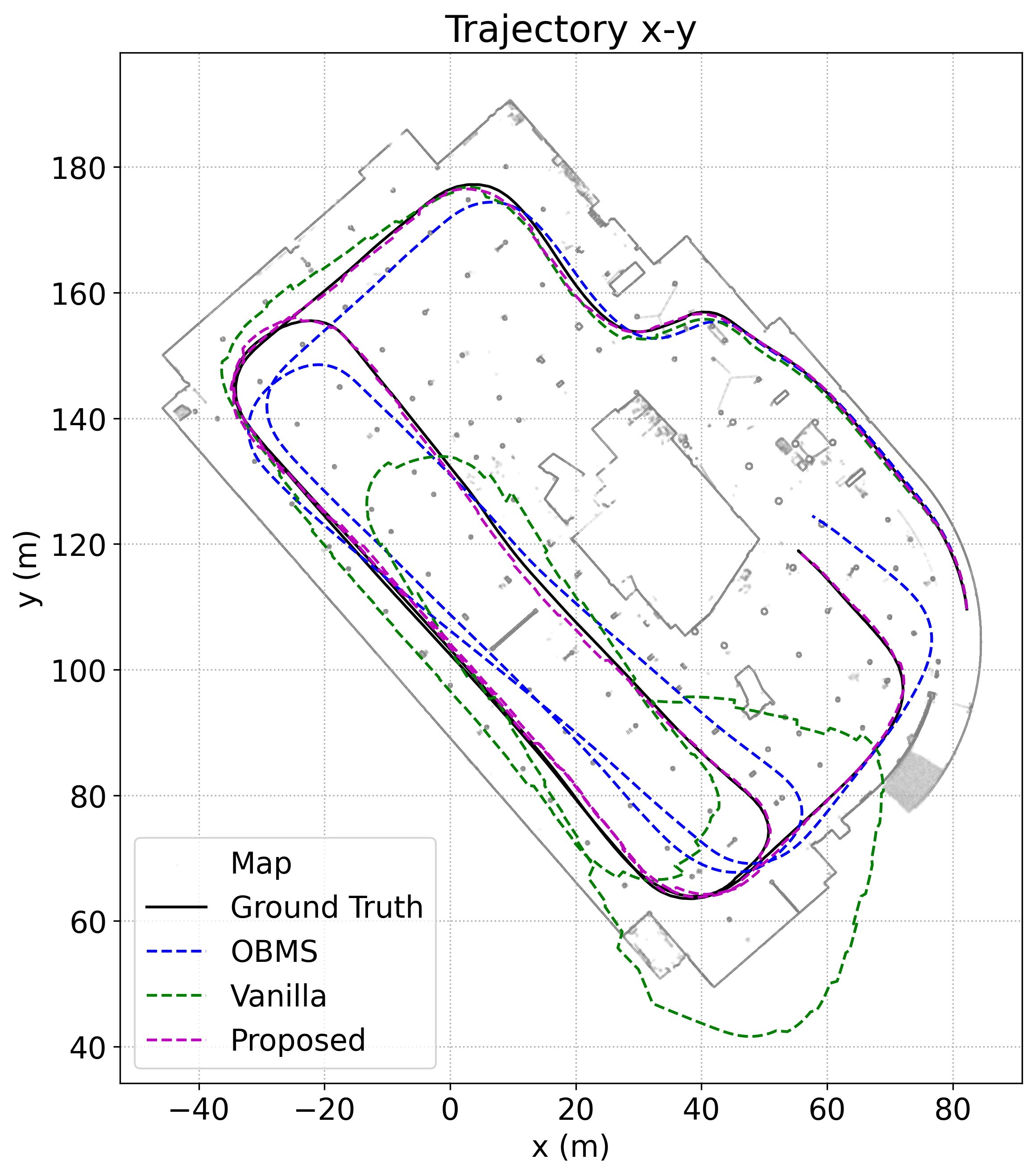}
  \label{fig:segment2_traj}
}
\caption{ Estimated trajectories of standalone OBMS, Vanilla VMR, and Proposed VMR in the indoor parking garage for (a) Segment I1 and (b) Segment I2.}
\label{fig:indoor_trajs}
\end{figure*}

\begin{figure*}
\centering
\subfloat[]{%
\includegraphics[height= 0.27 \linewidth]{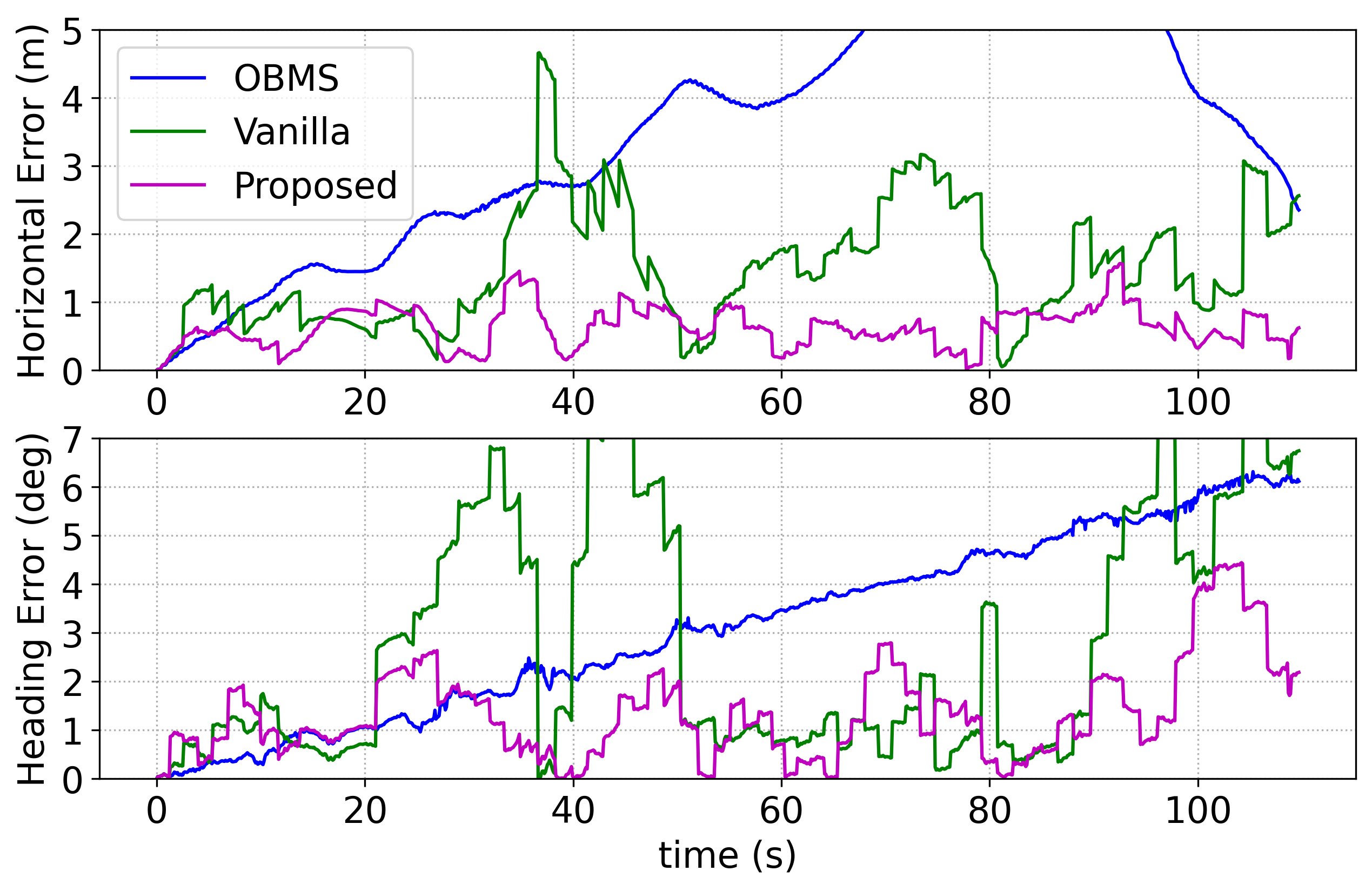}
  \label{fig:segment1_traj_errors}
}
\hspace{10pt}
\subfloat[]{%
\includegraphics[height= 0.27 \linewidth]{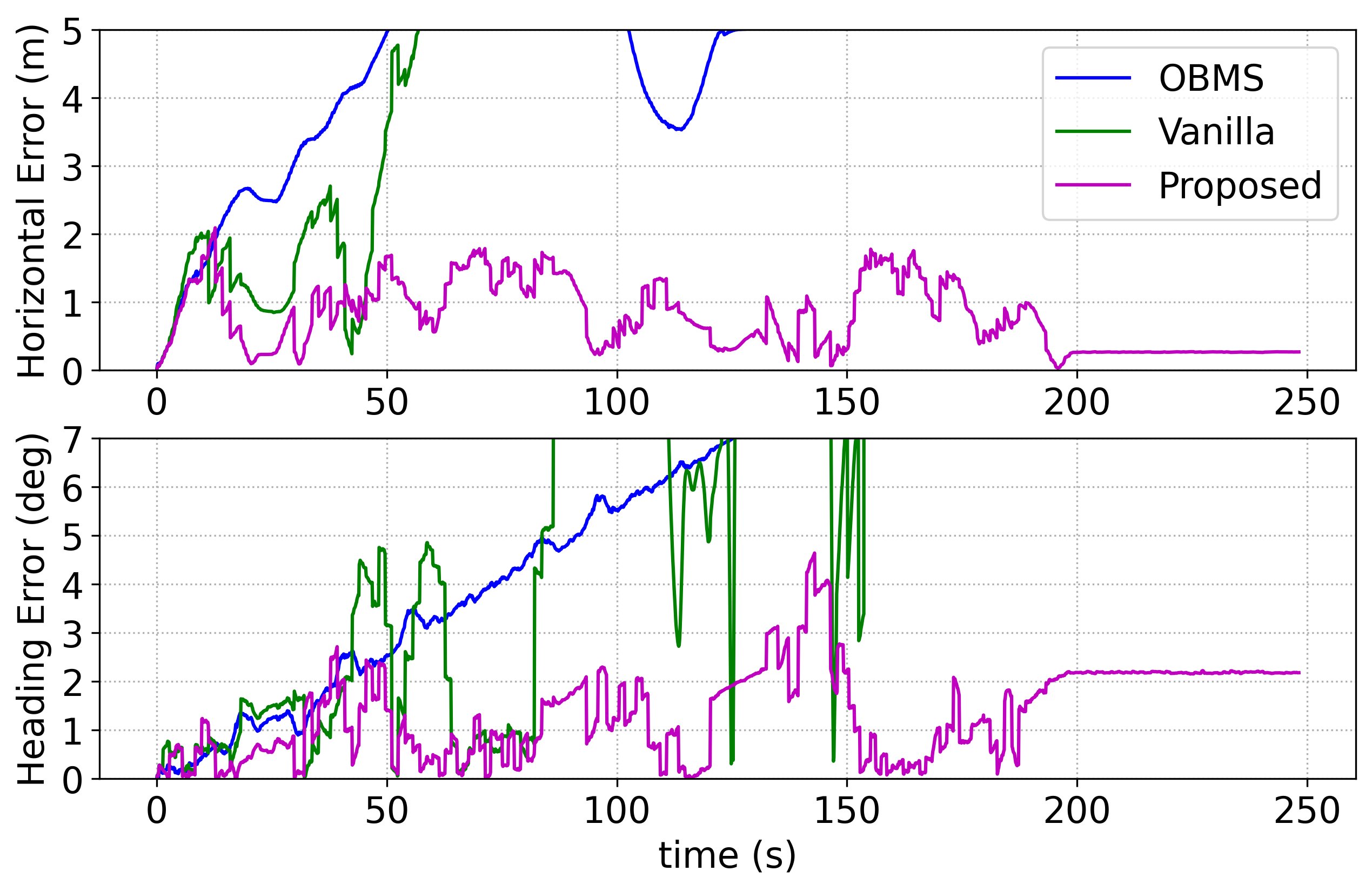}
  \label{fig:segment2_traj_errors}
}
\caption{ Horizontal and heading errors of standalone OBMS, Vanilla VMR, and Proposed VMR in the indoor parking garage for (a) Segment I1 and (b) Segment I2.}
\label{fig:indoor_traj_errors}
\end{figure*}

\begin{table}
\centering
\caption{Comparison of performance statistics for Indoor Segments 1 and 2.}
\resizebox{0.5\textwidth}{!}{%
{\small
\begin{tabular}{cc|ccc|ccc}
\toprule
\multicolumn{2}{c|}{\textbf{Metric}} & \multicolumn{3}{c|}{\textbf{Segment I1}} & \multicolumn{3}{c}{\textbf{Segment I2}} \\
\cmidrule(lr){3-5} \cmidrule(lr){6-8}
& & \textbf{OBMS} & \textbf{Vanilla} & \textbf{Proposed} & \textbf{OBMS} & \textbf{Vanilla} & \textbf{Proposed} \\
\midrule
\multicolumn{8}{c}{\textbf{Position}} \\
\midrule
\multirow{2}{*}{Horizontal} 
& RMSE $(\mathrm{m})$ \textdownarrow & $4.23$ & $1.74$ & $\mathbf{0.70}$ & $6.31$ & $20.45$ & $\mathbf{0.91}$ \\
& MaxAE $(\mathrm{m})$ \textdownarrow & $7.40$ & $4.66$ & $\mathbf{1.57}$ & $12.37$ & $35.10$ & $\mathbf{2.09}$ \\[5pt]

\multirow{2}{*}{Vertical} 
& RMSE $(\mathrm{m})$ \textdownarrow & $5.13$ & $0.43$ & $\mathbf{0.18}$ & $7.72$ & $0.45$ & $\mathbf{0.16}$ \\
& MaxAE $(\mathrm{m})$ \textdownarrow & $8.05$ & $0.68$ & $\mathbf{0.37}$ & $10.11$ & $0.77$ & $\mathbf{0.42}$ \\[5pt]

Lane-level &  $(\% < 1.5\,\mathrm{m})$  \textuparrow & $17.23$ & $57.16$ & $\mathbf{99.18}$ & $3.99$ & $10.96$ & $\mathbf{89.73}$ \\[5pt]
Submeter-level & $(\% < 1\,\mathrm{m})$ \textuparrow & $8.48$ & $34.73$ & $\mathbf{92.16}$ & $2.18$ & $6.69$ & $\mathbf{68.20}$ \\[5pt]

\midrule
\multicolumn{8}{c}{$\mathbf{Attitude}$} \\
\midrule

\multirow{2}{*}{Pitch} 
& RMSE $(^\circ)$ \textdownarrow & $1.50$ & $0.40$ & $\mathbf{0.39}$ & $1.98$ & $0.79$ & $\mathbf{0.40}$ \\
& MaxAE $(^\circ)$ \textdownarrow & $3.26$ & $1.49$ & $\mathbf{1.05}$ & $4.42$ & $3.20$ & $\mathbf{1.05}$ \\[5pt]

\multirow{2}{*}{Roll} 
& RMSE $(^\circ)$ \textdownarrow & $1.40$ & $0.97$ & $\mathbf{0.88}$ & $3.03$ & $1.16$ & $\mathbf{0.63}$ \\
& MaxAE $(^\circ)$ \textdownarrow & $3.07$ & $2.89$ & $\mathbf{1.95}$ & $7.52$ & $3.52$ & $\mathbf{1.79}$ \\[5pt]

\multirow{2}{*}{Heading} 
& RMSE $(^\circ)$ \textdownarrow & $3.63$ & $3.82$ & $\mathbf{1.67}$ & $8.02$ & $53.16$ & $\mathbf{1.64}$ \\
& MaxAE $(^\circ)$ \textdownarrow & $6.31$ & $9.39$ & $\mathbf{4.44}$ & $13.64$ & $167.72$ & $\mathbf{4.64}$ \\

\bottomrule
\end{tabular}%
}
}
\label{table:indoor-performance-combined-new}
\end{table}

\figurename~\ref{fig:indoor_trajs} illustrates a qualitative comparison between the three positioning solutions, where in both cases the proposed VMR system consistently outperforms the baseline OBMS and vanilla VMR approaches, providing a closer resemblance to the ground truth trajectory. These differences can be better demonstrated in the corresponding horizontal and heading error curves in \figurename~\ref{fig:indoor_traj_errors}.

A detailed quantitative comparison is summarized in \tablename~\ref{table:indoor-performance-combined-new}. In terms of horizontal accuracy, the OBMS solution accumulates drift over time due to its reliance on inertial propagation, resulting in substantial horizontal RMSE $4.23$\,m in segment I1 and $6.31$\,m in segment I2. Vanilla VMR provides modest improvements in segment I1 ($1.74$\,m RMSE) but completely fails in segment I2 ($20.45$\,m RMSE), where accumulated error exceeds the convergence threshold of the \ac{gicp} algorithm, leading to failed registrations and continuous drift. In contrast, the proposed VMR maintains robust performance with horizontal RMSEs of $0.70$\,m in segment I1 and $0.91$\,m in segment I2, achieving approximately $84.5$\% improvement over OBMS and $77.7$\% over vanilla across both segments.

Similar improvements are also evident in the table with respect to other navigation states.
For example, the vertical RMSE drops from $5.13$\,m (OBMS) and $0.43$\,m (vanilla) to $0.18$\,m with the proposed method, in segment I1. While in segment I2, vertical error dropped from $7.72$\,m for OBMS and $0.45$\,m for vanilla to $0.16$\,m. These results correspond to an average vertical RMSE improvement of $97.2$\% over OBMS and $61.3$\% over vanilla. Moreover, the heading RMSE remains consistent across both evaluations, $1.67$$^\circ$ and $1.64$$^\circ$, showing about $66$\% improvement over OBMS and $76.6$\% over vanilla.

Finally, the proposed method achieves $99.18$\% of position errors within $1.5$\,m (lane-level accuracy) and $92.16$\% within $1$\,m (submeter-level) in segment I1. In the more challenging segment I2, it maintains $89.73$\% of errors within $1.5$\,m and $68.20$\% within $1$\,m. This corresponds to an average improvement of $74.9$\% over OBMS and $59.5$\% over vanilla in submeter-level positioning accuracy.

\subsection{Outdoors}

\begin{figure*}
\centering
\subfloat[]{%
 \includegraphics[height= 0.27 \linewidth]{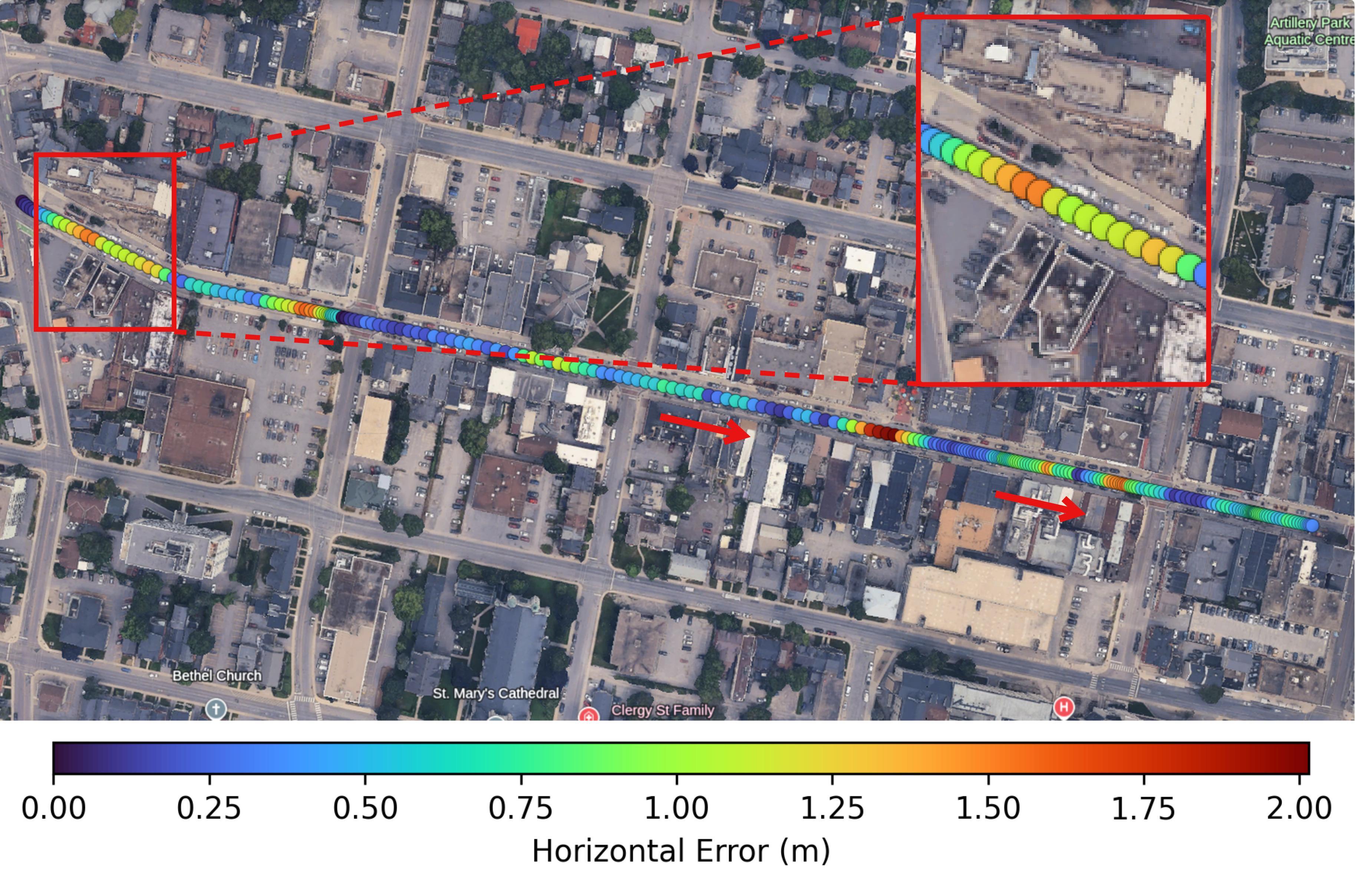}
 \label{fig:urban01_traj}
} 
\hspace{10pt}
\subfloat[]{%
\includegraphics[height= 0.27\linewidth]{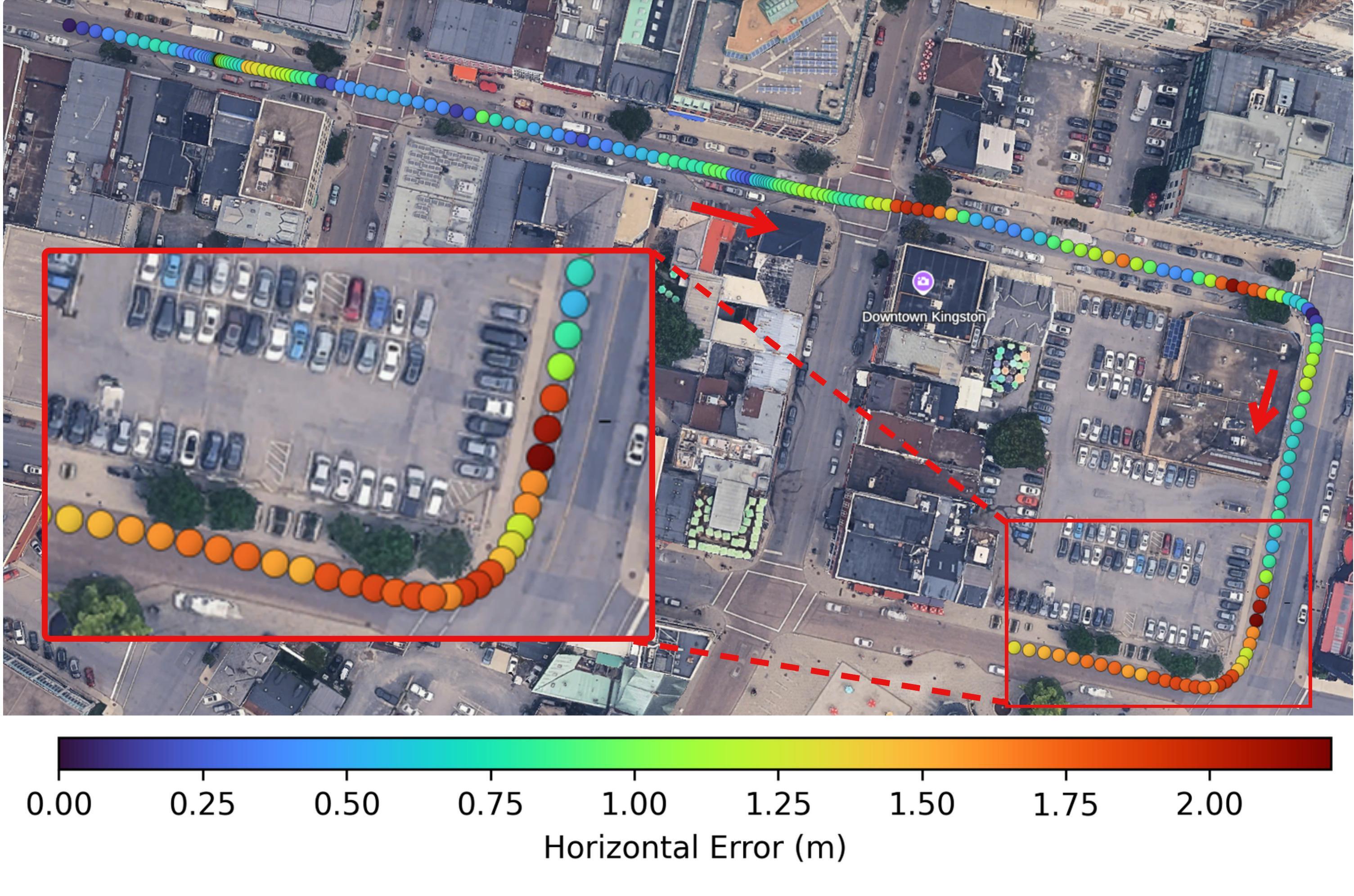}
  \label{fig:urban02_traj}
}
\\
\subfloat[]{%
 \includegraphics[height= 0.27 \linewidth]{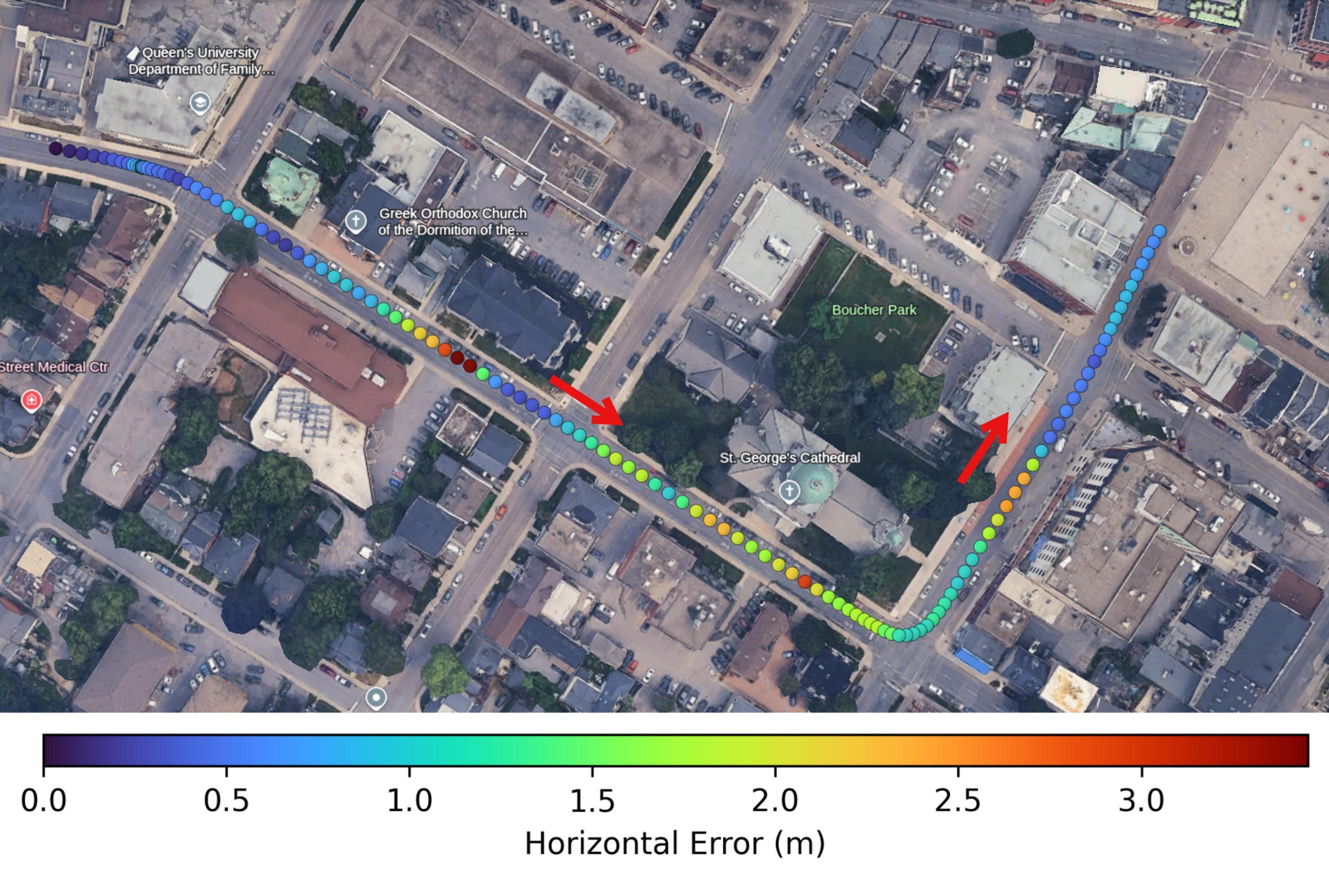}
 \label{fig:urban03_traj}
} 
\hspace{10pt}
\subfloat[]{%
\includegraphics[height= 0.27 \linewidth]{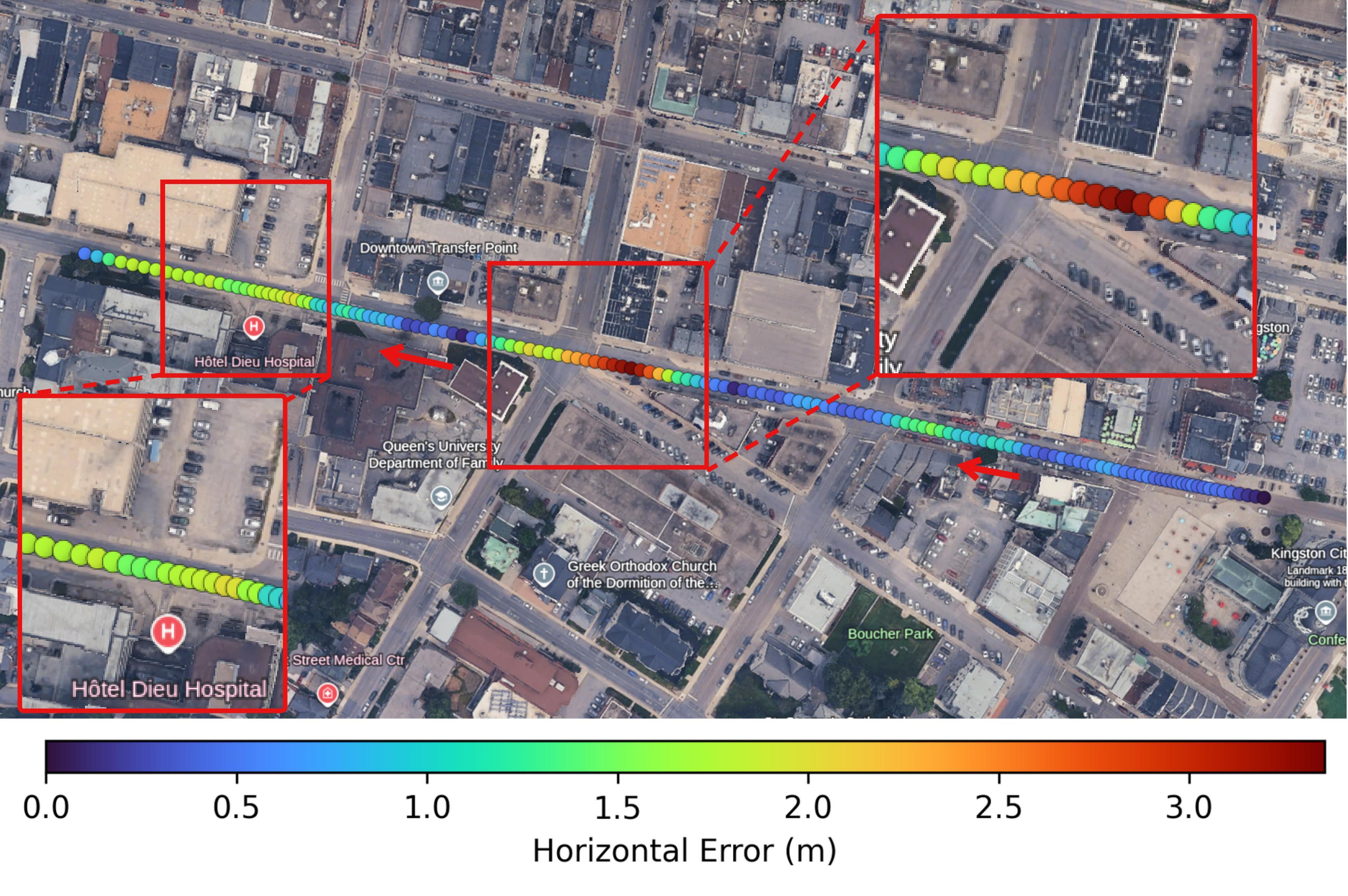}
  \label{fig:urban04_traj}
}
\caption{ Proposed VMR trajectory color-coded by horizontal positioning error for (a) Segment O1, (b) Segment O2, (c) Segment O3, and (d) Segment O4. }
\label{fig:outdoor_trajs}
\end{figure*}

\begin{figure*}
\centering
\subfloat[]{%
\includegraphics[height= 0.27 \linewidth]{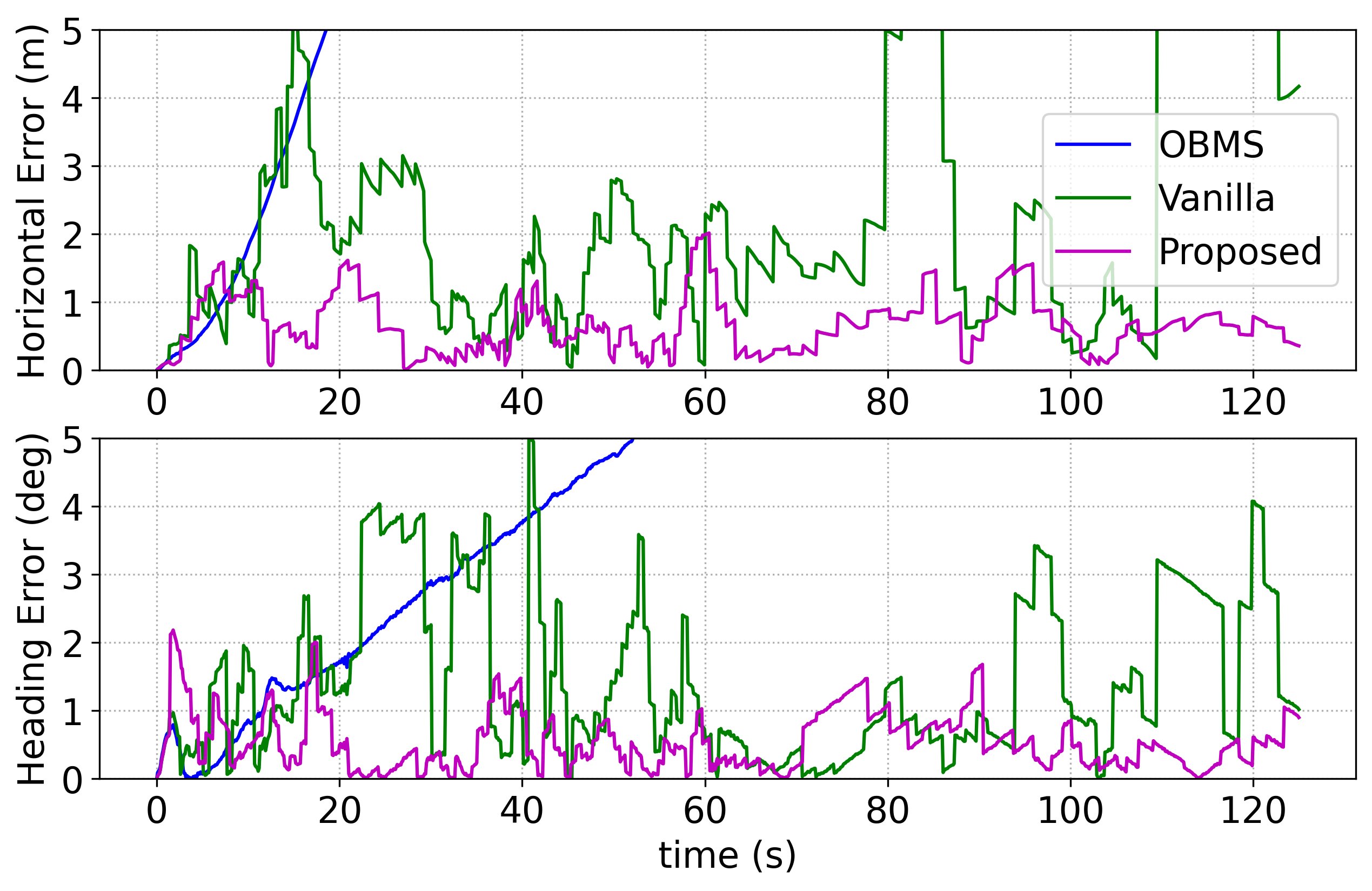}
  \label{fig:urban01_errors}
}
\hspace{10pt}
\subfloat[]{%
\includegraphics[height= 0.27 \linewidth]{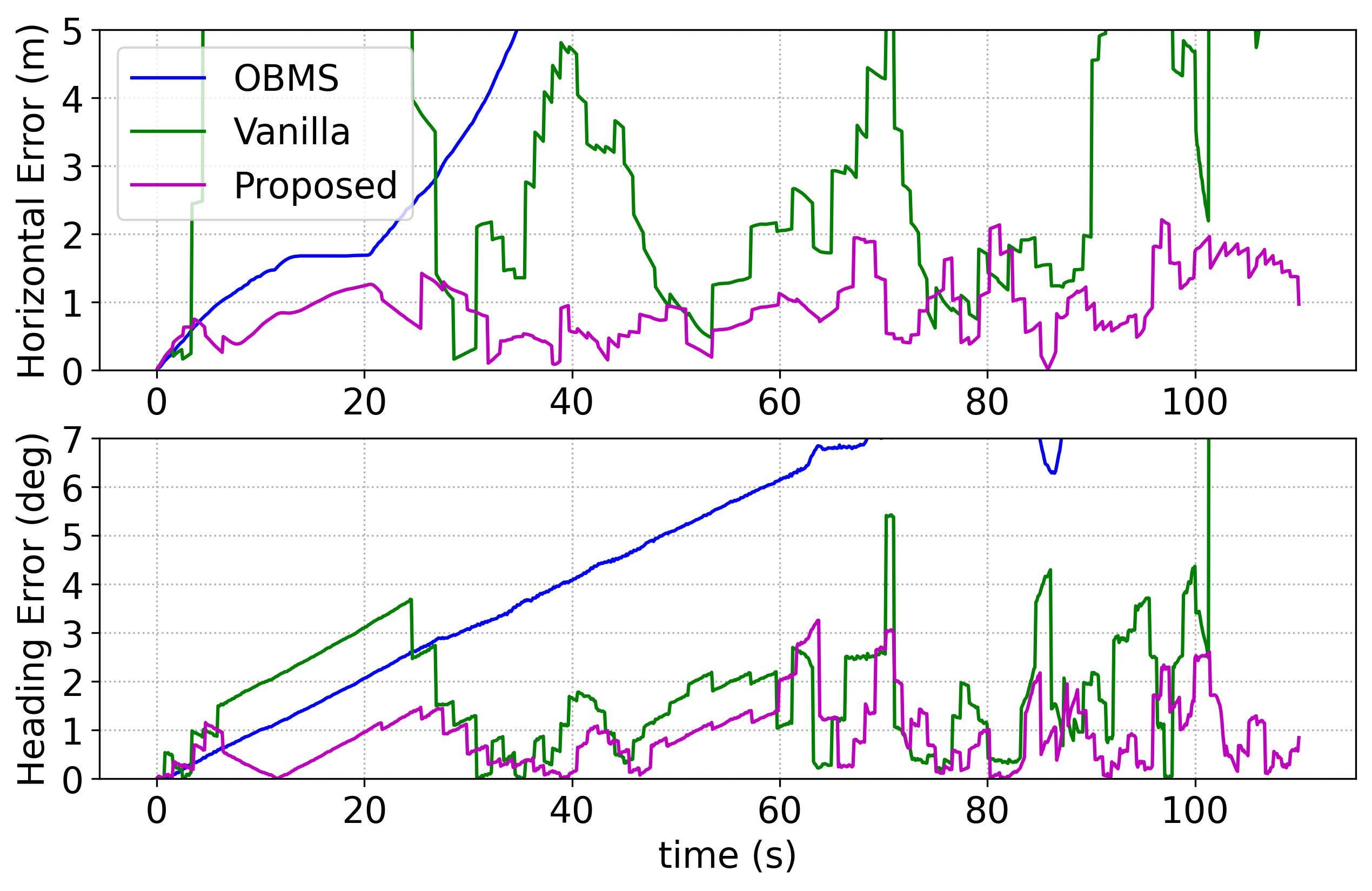}
  \label{fig:urban02_errors}
}
\\
\subfloat[]{%
 \includegraphics[height= 0.27 \linewidth]{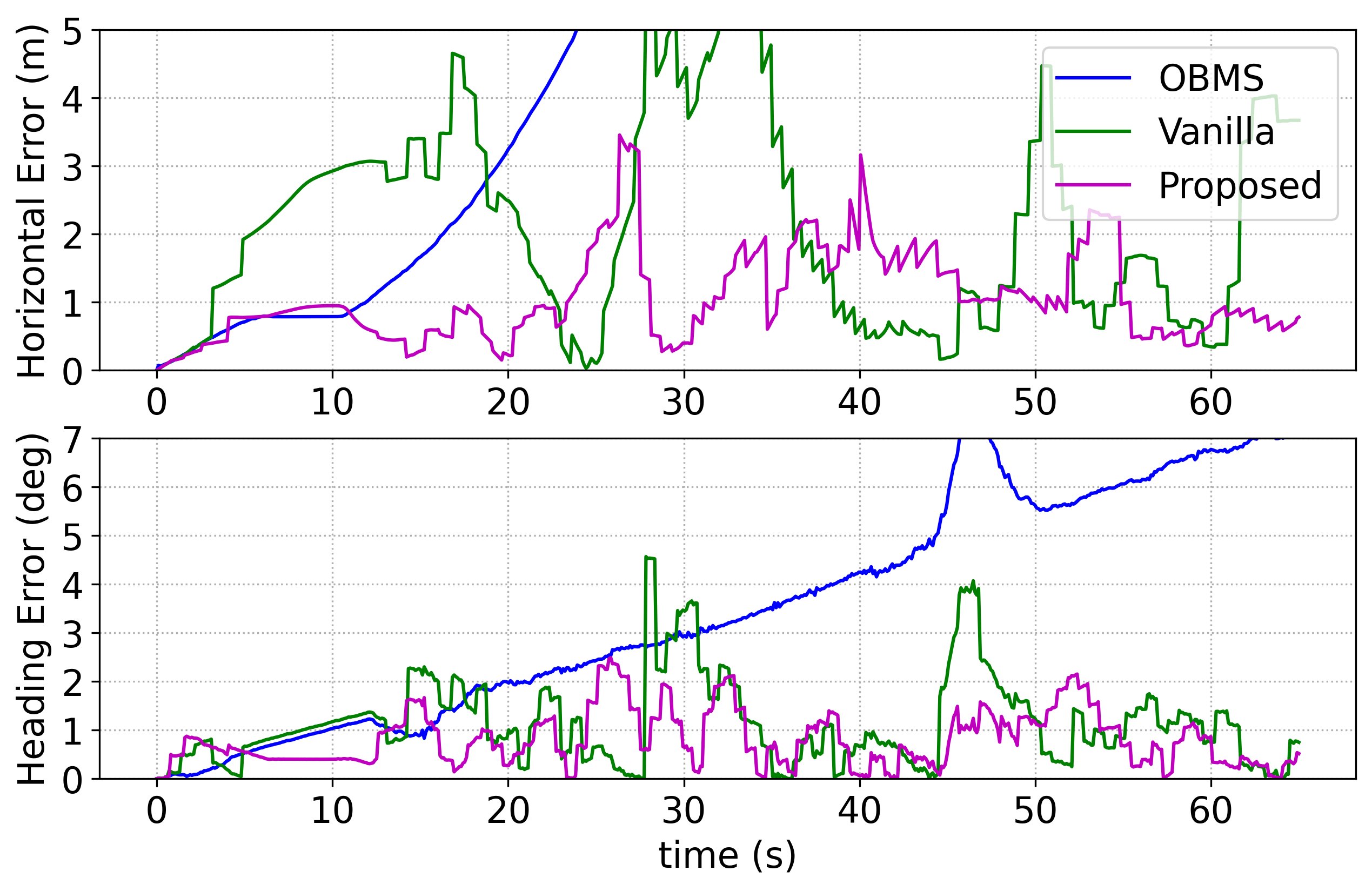}
 \label{fig:urban03_errors}
} 
\hspace{10pt}
\subfloat[]{%
\includegraphics[height= 0.27 \linewidth]{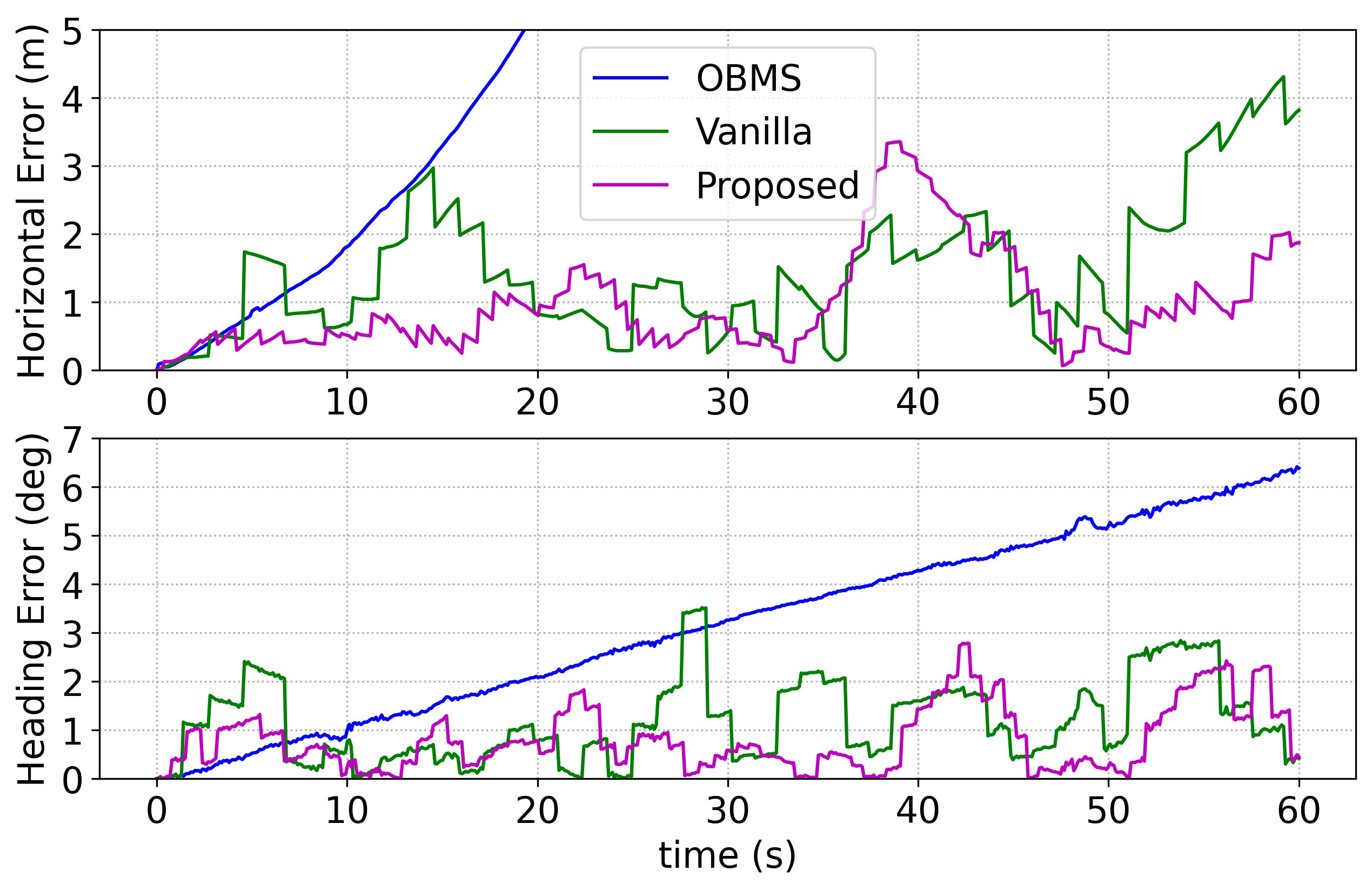}
  \label{fig:urban04_errors}
}
\caption{ Horizontal and heading errors for standalone OBMS, Vanilla VMR, and Proposed VMR in (a) Segment O1, (b) Segment O2, (c) Segment O3, and (d) Segment O4.}
\label{fig:outdoor_errors}
\end{figure*}

\begin{table*}
\centering
\caption{Comparison of performance statistics for Outdoor Segments 1–4.}
\resizebox{\textwidth}{!}{%
{\small
\begin{tabular}{cc|ccc|ccc|ccc|ccc}
\toprule
\multicolumn{2}{c|}{$\mathbf{Metric}$} 
& \multicolumn{3}{c|}{$\mathbf{Segment\ O1}$} 
& \multicolumn{3}{c|}{$\mathbf{Segment\ O2}$} 
& \multicolumn{3}{c|}{$\mathbf{Segment\ O3}$} 
& \multicolumn{3}{c}{$\mathbf{Segment\ O4}$} \\
\cmidrule(lr){3-5} \cmidrule(lr){6-8} \cmidrule(lr){9-11} \cmidrule(lr){12-14}
& & $\mathbf{OBMS}$ & $\mathbf{Vanilla}$ & $\mathbf{Proposed}$
  & $\mathbf{OBMS}$ & $\mathbf{Vanilla}$ & $\mathbf{Proposed}$
  & $\mathbf{OBMS}$ & $\mathbf{Vanilla}$ & $\mathbf{Proposed}$
  & $\mathbf{OBMS}$ & $\mathbf{Vanilla}$ & $\mathbf{Proposed}$ \\
\midrule
\multicolumn{14}{c}{$\mathbf{Position}$} \\
\midrule
\multirow{2}{*}{Horizontal} 
& RMSE $(\mathrm{m})$ \textdownarrow & $41.00$ & $3.39$ & $\mathbf{0.78}$ & $21.03$ & $6.15$ & $\mathbf{1.03}$ & $16.24$ & $2.67$ & $\mathbf{1.25}$ & $21.69$ & $1.82$ & $\mathbf{1.29}$ \\
& MaxAE $(\mathrm{m})$ \textdownarrow & $74.68$ & $9.51$ & $\mathbf{2.02}$ & $36.51$ & $26.79$ & $\mathbf{2.21}$ & $31.54$ & $6.56$ & $\mathbf{3.46}$ & $46.53$ & $3.67$ & $\mathbf{3.35}$ \\[5pt]

\multirow{2}{*}{Vertical} 
& RMSE $(\mathrm{m})$ \textdownarrow & $39.08$ & $0.94$ & $\mathbf{0.12}$ & $17.40$ & $0.97$ & $\mathbf{0.23}$ & $11.70$ & $1.02$ & $\mathbf{0.26}$ & $15.93$ & $1.03$ & $\mathbf{0.11}$ \\
& MaxAE $(\mathrm{m})$ \textdownarrow & $72.47$ & $1.47$ & $\mathbf{0.30}$ & $30.94$ & $1.63$ & $\mathbf{0.48}$ & $23.00$ & $1.53$ & $\mathbf{0.52}$ & $37.55$ & $1.37$ & $\mathbf{0.30}$ \\[5pt]

Lane-level & \% $<1.5\,\mathrm{m}$  \textuparrow & $7.43$ & $41.20$ & $\mathbf{95.20}$ & $10.52$ & $26.45$ & $\mathbf{85.91}$ & $20.57$ & $43.54$ & $\mathbf{76.00}$ & $12.14$ & $51.17$ & $\mathbf{72.43}$ \\[5pt]
Submeter-level & \% $<1\,\mathrm{m}$ \textuparrow & $5.84$ & $27.84$ & $\mathbf{81.68}$ & $5.62$ & $11.64$ & $\mathbf{62.73}$ & $17.00$ & $30.15$ & $\mathbf{60.15}$ & $8.29$ & $28.83$ & $\mathbf{57.86}$ \\[5pt]

\midrule
\multicolumn{14}{c}{$\mathbf{Attitude}$} \\
\midrule

\multirow{2}{*}{Pitch} 
& RMSE $(^\circ)$ \textdownarrow & $8.65$ & $0.78$ & $\mathbf{0.24}$ & $5.90$ & $0.90$ & $\mathbf{0.37}$ & $3.07$ & $0.72$ & $\mathbf{0.48}$ & $4.54$ & $0.40$ & $\mathbf{0.30}$ \\
& MaxAE $(^\circ)$ \textdownarrow & $14.66$ & $2.69$ & $\mathbf{1.00}$ & $10.22$ & $3.38$ & $\mathbf{1.02}$ & $5.62$ & $2.50$ & $\mathbf{2.34}$ & $8.68$ & $1.52$ & $\mathbf{1.32}$ \\[5pt]

\multirow{2}{*}{Roll} 
& RMSE $(^\circ)$ \textdownarrow & $0.67$ & $1.30$ & $\mathbf{0.64}$ & $3.81$ & $2.09$ & $\mathbf{1.25}$ & $2.62$ & $0.93$ & $\mathbf{0.68}$ & $1.45$ & $1.40$ & $\mathbf{0.76}$ \\
& MaxAE $(^\circ)$ \textdownarrow & $1.22$ & $3.57$ & $\mathbf{2.09}$ & $10.19$ & $5.65$ & $\mathbf{3.78}$ & $5.06$ & $2.42$ & $\mathbf{2.04}$ & $2.66$ & $4.40$ & $\mathbf{1.93}$ \\[5pt]

\multirow{2}{*}{Heading} 
& RMSE $(^\circ)$ \textdownarrow & $7.06$ & $1.85$ & $\mathbf{0.69}$ & $6.42$ & $8.67$ & $\mathbf{1.11}$ & $4.19$ & $1.44$ & $\mathbf{0.98}$ & $3.67$ & $1.66$ & $\mathbf{1.18}$ \\
& MaxAE $(^\circ)$ \textdownarrow & $12.24$ & $5.01$ & $\mathbf{2.18}$ & $11.55$ & $38.01$ & $\mathbf{3.25}$ & $7.32$ & $4.57$ & $\mathbf{2.49}$ & $6.41$ & $6.33$ & $\mathbf{3.45}$ \\
\bottomrule
\end{tabular}%
}
}
\label{table:outdoor-performance-combined-new}
\end{table*}

The proposed VMR solution was further evaluated across four outdoor segments varying in geometry and complexity, from straight urban roads to tight turns. Segment lengths ranged from $446$ to $700$\,m, durations between $70$ and $124$\,s, and speeds from $15$ to $36$\,km/h. Despite environmental variation such as open parks, parking lots, dense buildings, and intersections, the proposed method demonstrated reliable positioning performance.

The performance of the proposed VMR solution across all outdoor segments is illustrated in \figurename~\ref{fig:outdoor_trajs}, which shows Google Earth plots of the estimated trajectories color-coded by horizontal positioning error. The trajectory plots demonstrate a consistent, drift-free performance across varying outdoor conditions. \figurename~\ref{fig:outdoor_errors} further demonstrates the pose estimation performance across the same segments, highlighting the benefits of the proposed method in terms of horizontal position and heading errors compared to the examined baselines.

\tablename~\ref{table:outdoor-performance-combined-new} provides the quantitative analysis between the different segments and positioning solutions. The \ac{obms} baseline exhibited significant drift, with horizontal RMSEs ranging from $16.24$\,m to $41.00$\,m. Vanilla VMR improved this baseline by approximately $85$--$92$\%, but continued to suffer from poor convergence and failures, especially in feature-sparse areas. In contrast, the proposed method further reduced horizontal RMSE compared to vanilla VMR, resulting in final RMSEs between $0.78$\,m and $1.29$\,m. The proposed method achieves an average of $88$\% reduction over \ac{obms} and $74$\% over vanilla VMR in horizontal RMSE.

Submeter-level positioning varied based on scene geometry and feature availability. Segment O1 achieved the highest precision, with $81.68$\% of positioning errors within $1$\,m and $95.20$\% within $1.5$\,m. Segment O2, shown in \figurename~\ref{fig:urban02_traj} and involved sharp turns and open structures like parking lots, achieved $62.73$\% submeter accuracy and $85.91$\% within $1.5$\,m. Segments O3 and O4 performed similarly with $60.15$\% and $57.86$\% of positions within $1$\,m, respectively. These localized fluctuations in performance can be observed in the corresponding horizontal error plots in \figurename~\ref{fig:outdoor_errors}.

Heading estimation trends followed a similar pattern to horizontal errors. As shown in the error plots, both \ac{obms} and vanilla VMR suffered from increasing drift, with heading RMSEs ranging from $3.67^\circ$ to $8.67^\circ$. The proposed method, however, consistently reduced heading RMSE by approximately $70$--$90$\% relative to both baselines. Averaged across all segments, the heading RMSE improvement was around $80$\% over \ac{obms} and $75$\% over vanilla VMR, maintaining heading errors below $1.3^\circ$ in all cases. Segment O1 achieved the best result ($0.69^\circ$) due to its linear geometry and stable features, while segment O4 had the highest heading RMSE ($1.18^\circ$) as a result of visual aliasing and weak structural context near intersections.

Scene complexity varied across segments, causing localized increases in error for all methods. In Segment O1, errors rose at the beginning due to an open area with few features and near intersections where abrupt depth changes occurred (\figurename~\ref{fig:urban01_traj}, \figurename~\ref{fig:urban01_errors}). Segment O2 showed error spikes primarily after the second $90^\circ$ turn near open parking garages and parks across from it (\figurename~\ref{fig:urban02_traj}, \figurename~\ref{fig:urban02_errors}). Segment O3’s errors fluctuated along the route, notably when passing next to a small parking lot, otherwise it showed consistent errors (\figurename~\ref{fig:urban03_traj}, \figurename~\ref{fig:urban03_errors}). Segment O4 experienced increased errors at a central wide intersection with distant buildings and near a parking garage and hospital entrance (\figurename~\ref{fig:urban04_traj}, \figurename~\ref{fig:urban04_errors}). Despite these error increases, the robust fusion filter within the proposed system effectively leveraged INS, speed corrections, and dynamic tuning to recover positioning accuracy, reduce drift, and maintain submeter-level precision throughout the trajectories.

\section{Conclusion}\label{sec:conclusion}
This paper presented a robust, scalable positioning solution for autonomous vehicles operating in GNSS-challenged environments. Leveraging recent advances in monocular depth estimation, semantic filtering, and 3-D map alignment, the proposed system delivers consistent sub-meter accuracy using only low-cost sensors. Unlike traditional LiDAR systems, this approach is lightweight, cost-effective, and deployable in existing vehicle sensor architectures. 

Future directions include extending to multi-camera systems for wider field-of-view coverage and evaluating cross-city generalization under diverse lighting and seasonal conditions. This study lays the foundation for the practical deployment of vision-driven navigation systems in real-world autonomous platforms.

\section*{Acknowledgments}
The authors acknowledge the Centre for Advanced Computing (CAC), Queen’s University, and Micro Engineering Tech Inc. (METI) for providing the 3-D maps used in this research.


\begin{IEEEbiography}[{\includegraphics[width=1in,height=1.25in,clip,keepaspectratio]{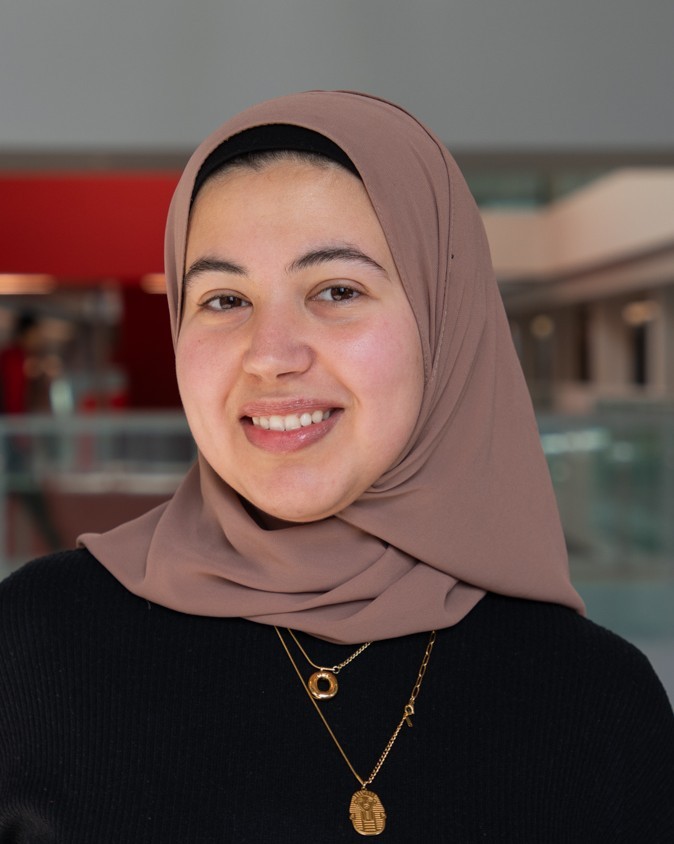}}]{Ola Elmaghraby} received the B.Sc. degree in Electrical Engineering from Cairo University, Egypt, in 2023. She received her M.Sc. degree in Electrical and Computer Engineering at Queen's University, Canada, in 2025. Ola is also a member of the Navigation and Instrumentation Research Laboratory (NavINST) at the Royal Military College of Canada (RMCC). Her research interests include positioning for autonomous vehicles using computer vision integrated with inertial and odometer sensors.
\end{IEEEbiography}

\begin{IEEEbiography}[{\includegraphics[width=1in,height=1.25in,clip,keepaspectratio]{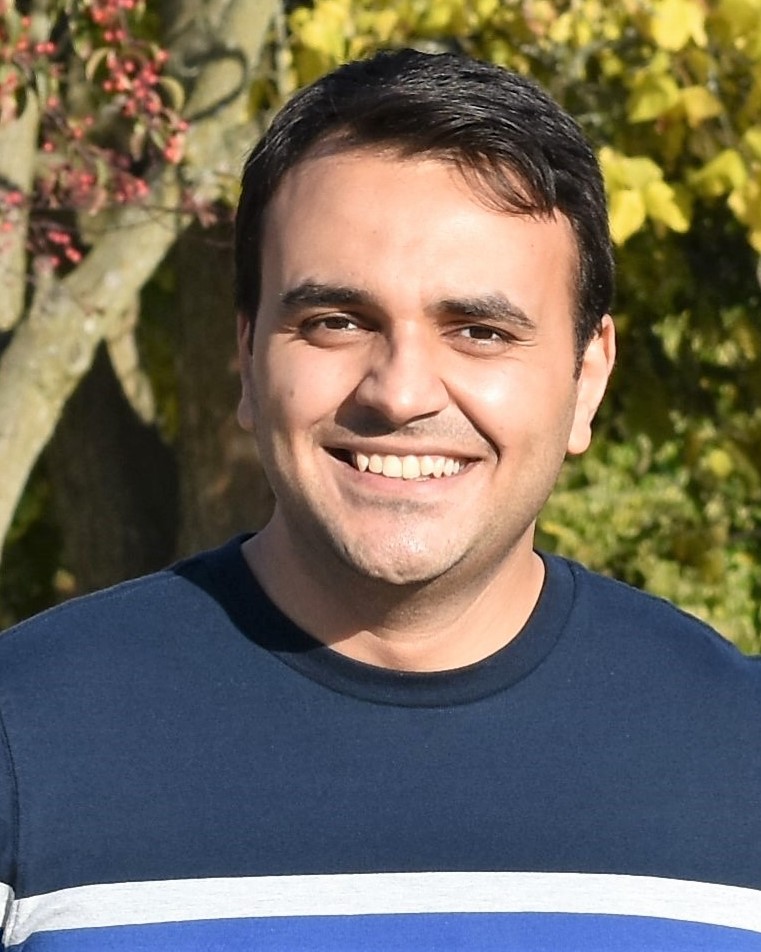}}]{Eslam Mounier} (M'25) received the B.Sc. and M.Sc. degrees in computer and systems engineering (CSE) from Ain Shams University, Cairo, Egypt, in 2014 and 2020, and the Ph.D. degree in electrical and computer engineering (ECE) from Queen’s University, Kingston, ON, Canada in 2025.
He is currently a Postdoctoral Researcher with the School of Computing, Queen’s University, Kingston, ON, Canada, and also with the Navigation and Instrumentation Research Laboratory (NavINST), Royal Military College of Canada (RMCC), Kingston. His primary research interests include multisensor navigation systems for autonomous vehicles, advanced machine perception and computer vision techniques, and machine learning and deep learning applications.
\end{IEEEbiography}

\begin{IEEEbiography}[{\includegraphics[width=1in,height=1.25in,clip,keepaspectratio]{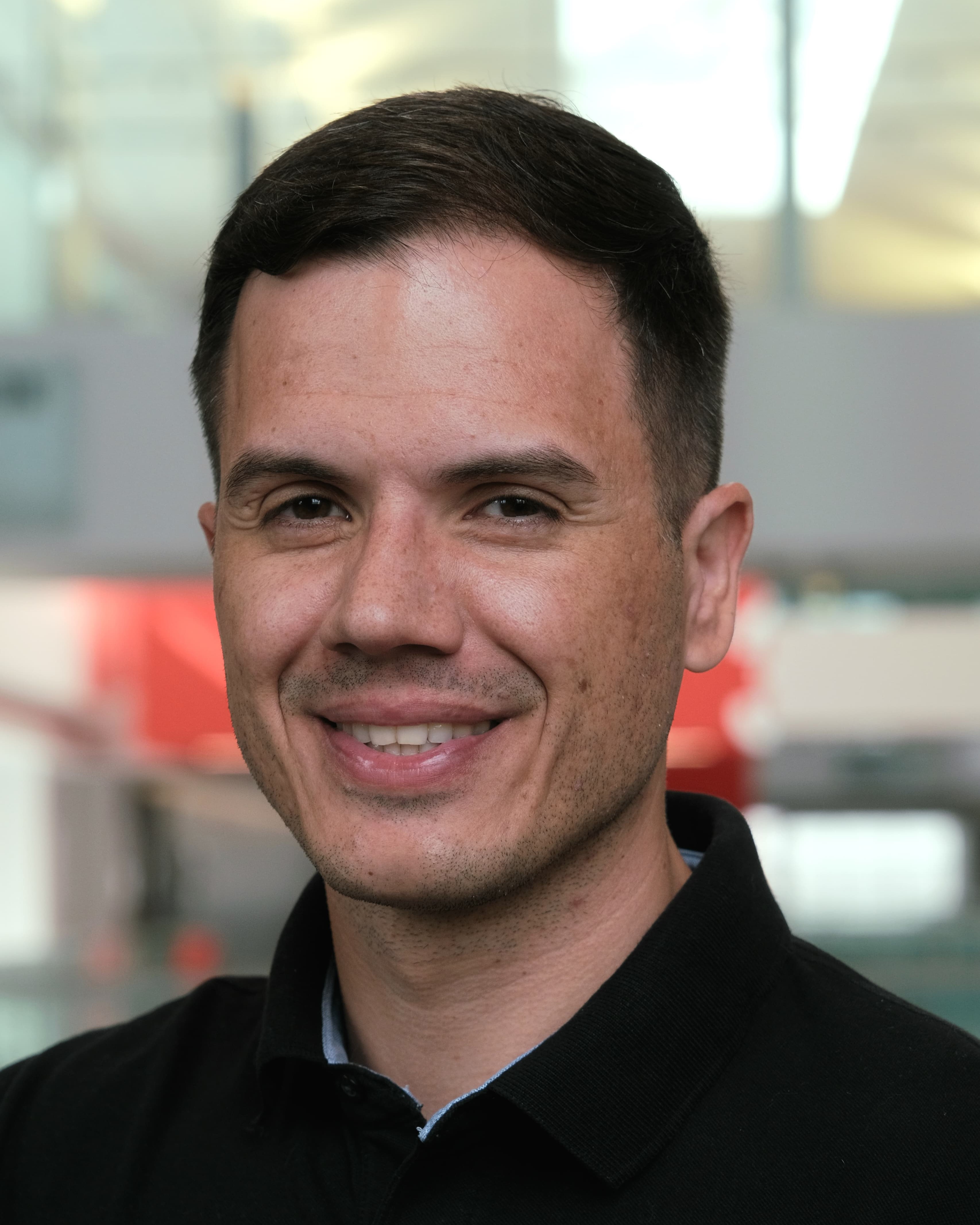}}]{Paulo Ricardo Marques de Araujo} (M'24) received a Ph.D. in Electrical and Computer Engineering from Queen’s University, Kingston, ON, Canada.
His broader research interests include autonomous systems, robotics, machine learning, and digital manufacturing.
\end{IEEEbiography}

\begin{IEEEbiography}[{\includegraphics[width=1in,height=1.25in,clip,keepaspectratio]{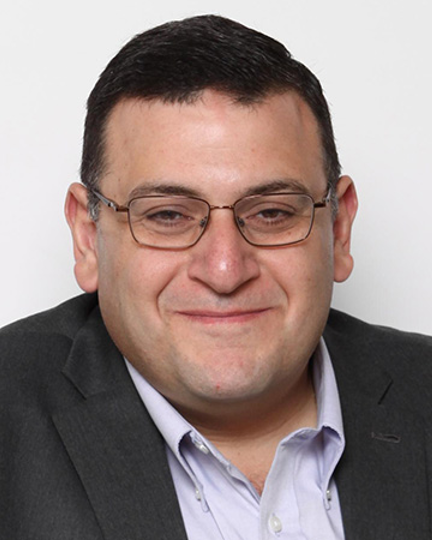}}]{Aboelmagd Noureldin} (SM'08) is a Professor and Canada Research Chair (CRC) at the Department of Electrical and Computer Engineering, Royal Military College of Canada (RMC), with Cross-Appointment at both the School of Computing and the Department of Electrical and Computer Engineering, Queen’s University. He is also the founding director of the Navigation and Instrumentation (NavINST) research group at RMC, a unique world-class research facility in GNSS, wireless positioning, inertial navigation, remote sensing, and multisensory fusion for navigation and guidance. Dr. Noureldin holds a Ph.D. in Electrical and Computer Engineering (2002) from The University of Calgary, Alberta, Canada. In addition, he has a B.Sc. in Electrical Engineering (1993) and an M.Sc. degree in Engineering Physics (1997), both from Cairo University, Egypt. Dr. Noureldin is a Senior member of the IEEE and a professional member of the Institute of Navigation (ION). He published two books, four book chapters, and over 350 papers in academic journals, conferences, and workshop proceedings, for which he received several awards. Dr. Noureddin’s research led to 13 patents and several technologies licensed to the industry in position, location, and navigation systems.
\end{IEEEbiography}

\vfill

\end{document}